\documentclass[10pt,twocolumn,letterpaper]{article}
\pdfsuppressptexinfo=-1 \pdfinfoomitdate=1 \pdftrailerid{}

\usepackage[pagenumbers]{cvpr}      %

\usepackage{cuted}    %
\usepackage{currfile} %
\usepackage{caption}  %

\usepackage{booktabs}
\usepackage{multirow}
\usepackage{array}
\usepackage{colortbl}
\usepackage{amsmath,amssymb}

\definecolor{trainedbg}{HTML}{F3ECDB}
\definecolor{freebg}{HTML}{E9F1E4}
\definecolor{bestbg}{HTML}{FFE04D}
\newcommand{\trained}{\rowcolor{trainedbg}}
\newcommand{\tfree}{\rowcolor{freebg}}
\newcommand{\best}[1]{\cellcolor{bestbg}#1}
\newcommand{\hlword}[1]{{\setlength{\fboxsep}{0.8pt}\colorbox{bestbg}{\strut #1}}}

\newcommand{\pp}[1]{$#1$}                                %
\newcommand{\ci}[2]{{\scriptsize[$#1$,\,$#2$]}}          %

\newcommand{\dsl}[1]{\texttt{\small #1}}                  %
\newcommand{\supref}[1]{Supp.~Sec.~\ref{#1}}

\newcommand{\method}{HeightProg}
\newcommand{\bench}{AltiBench}



\usepackage{tcolorbox}
\tcbuselibrary{breakable}
\newtcolorbox{promptbox}[1]{breakable, colback=black!3, colframe=black!45, colbacktitle=black!12,
  coltitle=black, fonttitle=\footnotesize\bfseries\sffamily, title={#1}, boxrule=0.5pt, arc=2pt, left=5pt, right=5pt,
  top=2pt, bottom=2pt, before upper=\footnotesize}

\definecolor{cvprblue}{rgb}{0.21,0.49,0.74}
\usepackage[pagebackref,breaklinks,colorlinks,allcolors=cvprblue]{hyperref}
\title{HeightProg: Typed Programs over Frozen Sensors\\for Metric 3D Question Answering on Aerial Images}

\usepackage{fontawesome5}
\AddToShipoutPicture*{\AtPageUpperLeft{\put(\LenToUnit{\dimexpr1in+\oddsidemargin\relax},\LenToUnit{-0.62in}){%
  \makebox[0pt][l]{\small\itshape\color{black!55}Preprint}}}}
\author{Yuhang Jiang$^{1}$\quad Guohui Deng$^{1}$\quad Zhongwen Hu$^{2}$\quad Nan Wang$^{3}$\quad Jian Ma$^{1}$\\[2pt]
$^{1}$Anhui University\quad $^{2}$Shenzhen University\quad $^{3}$Anhui Normal University\\[8pt]
{\normalsize\bfseries
\faGlobe~\href{https://avalon-s.github.io/HeightProg}{Project Page}\qquad
\raisebox{-0.35ex}{\includegraphics[height=2.4ex]{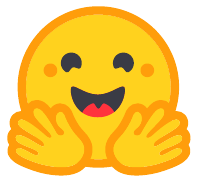}}~\href{https://huggingface.co/datasets/Avalon-S/AltiBench}{AltiBench}\qquad
\faGithub~\href{https://github.com/Avalon-S/HeightProg}{Code}}}

\begin{document}
\maketitle
\begin{strip}
\centering
\includegraphics[width=\linewidth]{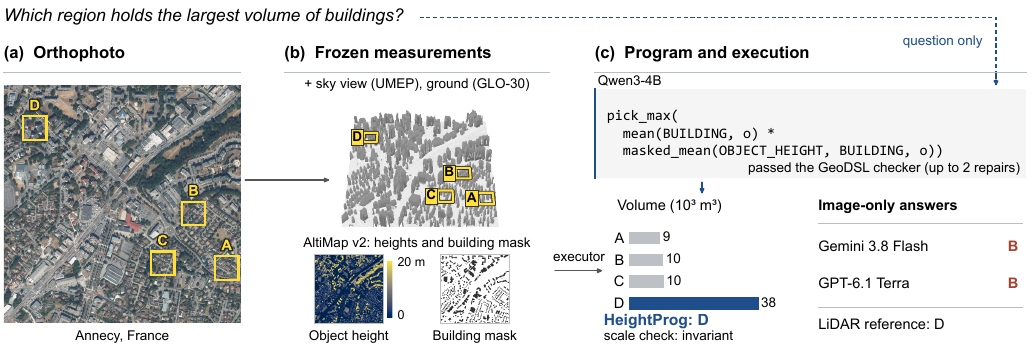}
\captionsetup{hypcap=false}
\captionof{figure}{\textbf{\method{} on a volume question from \bench{}.} The language model sees only the
question, and its typed program runs on metric layers from frozen sensors. GPT-6.1 Terra and Gemini 3.8 Flash judge from appearance and choose
region B, a cluster of small houses, over D, whose two buildings are four times as tall.}
\captionsetup{hypcap=true}
\label{fig:\currfilebase}%
\end{strip}

\begin{abstract}
Many questions about aerial images ask for metric 3D quantities that the pixels do not show directly, such as the
height of buildings, which of several areas lies lowest or whether two points can see each other. Current methods let
vision--language models (VLMs) answer from the image. On Geo3DVQA they answer at most 57\% of the questions correctly, even after fine-tuning on
100K questions from the same region, and their errors cannot be traced to perception or to reasoning.
Monocular height estimators provide metric layers, but their scale error changes from tile to tile. We present
\method{}, a training-free system that separates measurement from reasoning. Two frozen sensors, a monocular height
estimator and a public 30\,m elevation model, turn the image into metric layers. A 4B language model that never sees
the image compiles each question into a typed program, which a deterministic executor runs on the layers. Borrowing
from dimensional analysis, the type system marks, before execution, the answers that do not depend on the height
scale. Because some existing benchmarks use answers written or judged by language models, which makes errors hard to
attribute, we build \bench{}, with 2,889 questions on 120 one-kilometre tiles in Switzerland and France. Each question
states how its quantity is defined, and a script computes the answers from airborne LiDAR and national topographic
maps. Given the measurements, GPT-6.1 Terra rises from 60.2\% to 74.2\% on \bench{}, so access to measurements is a
major bottleneck. \method{} reaches 74.6\% on Geo3DVQA, 25 points above the published fine-tuned model, and 72.6\% on
\bench{}, 10 points above the best image-only VLM and 13 to 16 points above free-form visual programs over the same
sensors.
\end{abstract}

\section{Introduction}
\label{sec:intro}

Many questions asked of aerial images are about the third dimension. A planner wants the mean height of the buildings
in a block, a solar installer the rooftop that sees the most sky, a hydrologist the lowest of four fields, and a
surveyor whether a mast is visible from a hilltop. Their answers are metric and depend on heights that an orthophoto
shows only through shadows, perspective cues and context. Vision--language models (VLMs) are now asked to answer such
questions directly from the image, and the benchmarks built for this purpose find them weak. On
Geo3DVQA~\cite{geo3dvqa}, GPT-4o and Gemini-2.5-Flash answer 29\% and 33\% of the short-answer questions, and
fine-tuning Qwen2.5-VL-7B~\cite{qwen25vl} on 100K questions from the same region raises this to 50\%. Height-aware
instruction tuning~\cite{geoheight} and spatially trained VLMs~\cite{spatialrgpt,spatialreasoner} pursue the same
end-to-end route.

We find that access to task-relevant geometric measurements is a major bottleneck for these models. On our benchmark \bench{},
open-weight VLMs with reasoning off stay within 15 points of a constant answer, and the frontier models Gemini 3.8 Flash and GPT-6.1
Terra reach 62 and 60\%. When we give the same GPT model the heights, land cover, sky view and ground elevation our sensors measure
in each region, it rises to 74\%. Heights and elevations alone give under a quarter of this gain, and derived quantities such as counts, slopes and
clearances give the rest (Tab.~\ref{tab:main}). In
Fig.~\ref{fig:teaser}, a frontier model describes the buildings plausibly and still picks the wrong region.

We therefore propose \method{}, which separates measurement from reasoning. Frozen sensors turn the image into metric
layers of object height, land cover, sky view and bare ground. A 4B language model reads the question but not the
image and writes one program in GeoDSL, a small typed language over these layers, and a deterministic executor runs
it. One factor per tile removes half or more of a monocular estimator's squared height error, and before execution
we use the program's types to identify answers invariant to such a stretch.

We also introduce \bench{}, built from airborne LiDAR and national topographic maps in two countries. It holds
2,889 test questions of 15 types on 120 one-kilometre tiles in six Swiss and six French areas, from city centres to
Alpine valleys. A script that shares no code with
\method{} computes the answers from LiDAR heights and mapped building footprints, and the same generator produces
training and development splits from disjoint tiles.

Without any training, \method{} answers 72.6\% of the \bench{} questions, 10 points more than the best image-only
VLM and 31 points more than in-domain fine-tuning. On Geo3DVQA
it reaches 74.6\%, 25 points above the published fine-tuned model. It comes within 1.7 points of GPT-6.1 Terra reading the same
measurements, draws level with it when a larger text-only model writes the programs, and names in each program the
layers it reads. The same
design carries over to street-level images of SpatialRGPT-Bench~\cite{spatialrgpt} with a monocular depth model as the
sensor. Our contributions are as follows.
\begin{itemize}
\item \textbf{Diagnosis and benchmark.} We show that the lack of task-relevant geometric measurements limits VLMs on
  metric 3D questions about aerial images, and we release \bench{}, a two-country benchmark whose answers come from LiDAR and topographic maps,
  together with its generator and evaluation code.
\item \textbf{\method{}.} A training-free system in which frozen sensors measure and a 4B text-only model reasons in
  typed programs. With the same writer and sensors, it leads free-form visual programs on \bench{} by 13 to 16 points and its own 4B model reading the measurements as text by 10.
\item \textbf{Scale-invariance check and estimator testbed.} A static rule, much as in dimensional analysis, marks the
  answers that no tile-wide rescaling of the height sensor can change, and its degrees expose 98 programs that answer
  in metres without reading a height. \bench{} ranks height estimators by their use in reasoning, even without answer labels.
\end{itemize}

\section{Related Work}
\label{sec:related}

\noindent\textbf{3D Questions on Remote Sensing Images.}
Remote sensing VQA began with presence, counting and comparison questions answered from map data~\cite{rsvqa} and has
moved to height. Geo3DVQA~\cite{geo3dvqa} asks 16 types of questions about elevation, sky view factor and land cover on
GeoNRW tiles~\cite{geonrw}. GeoHeightChat, introduced with GeoHeight-Bench and its terrain extension~\cite{geoheight}, is a height-aware VLM trained on generated instructions, and
VertiCue-Bench~\cite{verticue} compares how well multimodal models use height evidence shown in the image,
returned by a tool or given as text. Remote sensing VLMs such as GeoChat~\cite{geochat}
and EarthDial~\cite{earthdial} are trained on large instruction sets but not on metric heights. Most of these answer
from the image end to end, and the strongest published Geo3DVQA result fine-tunes on the test region.

\noindent\textbf{Spatial Reasoning in VLMs.}
For ground-level images, SpatialVLM~\cite{spatialvlm} and SpatialRGPT~\cite{spatialrgpt} synthesise metric spatial
questions with depth estimators and fine-tune VLMs on them, SpatialReasoner-R1~\cite{spatialreasoner} adds preference optimisation, and
SpaceThinker~\cite{spacethinker} fine-tunes on synthetic reasoning traces. These models compute
distances inside the network, even where SpatialRGPT takes depth as an extra input. \method{} keeps the arithmetic in an external executor,
which lets it swap the sensor and analyse its errors without retraining.

\noindent\textbf{Visual Programming.}
Neural module networks~\cite{nmn,iep} execute programs predicted from the question with modules trained on the task,
and NS-VQA~\cite{nsvqa} runs such programs with deterministic modules over a parsed scene. VisProg~\cite{visprog} and ViperGPT~\cite{vipergpt} let a language model compose pretrained vision modules into a program, and
later work adds tools, sketches and dynamic code~\cite{sketchpad,pyvision}. For 3D questions on ground-level images,
VADAR~\cite{vadar} and pySpatial~\cite{pyspatial} write Python over depth, detection and reconstruction tools. In remote sensing, GeoSelect~\cite{geoselect}
compiles referring expressions with a frozen text-only language model into a typed spatial language that a checker
accepts and a deterministic executor runs, and GeoProg3D~\cite{geoprog3d} composes queries over city-scale
3D language fields built from many views. QASR, introduced with SQuID~\cite{squid}, writes programs
that segment satellite images and measure 2D distances and areas, and reference masks raise its accuracy from 42 to
83\%. Tool-calling agents for Earth observation~\cite{earthagent,thinkgeo,dora}
and the concurrent M3SunAgent~\cite{m3sunagent} plan tool calls over imagery, elevation layers or depth. \method{} applies the design of GeoSelect
to metric 3D questions over height and terrain layers, and the degrees in height of its types resemble dimension
types~\cite{kennedy}. We compare with VisProg- and ViperGPT-style programs over our sensors.

\noindent\textbf{Monocular Height and Depth Estimation.}
Height estimation from a single aerial image has progressed from regression networks~\cite{im2height} to models trained
on synthetic and multi-city data~\cite{synrs3d} and to canopy height maps of large regions~\cite{hrch}. Metric depth models such
as Depth Pro~\cite{depthpro} do the same for ground-level images. Their errors are reported per pixel, and we
measure them by their effect on answers.

\section{\method{}}
\label{sec:method}

A question comes with one orthophoto $I$ of known ground sampling distance and refers to its options by image
coordinates, either as labelled regions, as points or as one target region. The answer is a number with a unit, one
option, an ordering of options or a yes or no. \method{} answers in three stages (Fig.~\ref{fig:teaser}). Frozen
sensors turn $I$ into metric layers (Sec.~\ref{sec:sensors}), a text-only language model compiles the question into a
typed program (Sec.~\ref{sec:geodsl}), and a deterministic executor runs the program on the layers. The option geometry
is parsed from the question text by fixed rules, so the program only states what to compute for an option $o$.

\subsection{Metric Sensors}
\label{sec:sensors}

\noindent\textbf{Objects.}
A monocular height estimator maps $I$ to an object height map $\hat h$, the height of buildings and trees above the
ground in metres. We use the public AltiMap v2 checkpoint~\cite{altimap}, an RS3DAda model~\cite{synrs3d} trained on aerial
images of three US cities~\cite{gamus} and on synthetic scenes. Its segmentation head provides buildings, trees and other natural cover. We compute sky view factors from the predicted object heights over flat ground with
UMEP~\cite{svfcast,umep} over 153 sky patches.

\noindent\textbf{Ground.}
A single image does not show absolute elevation, and a monocular estimator predicts heights relative to an unseen
ground. We add the Copernicus GLO-30 digital elevation model~\cite{copdem}, a global 30\,m surface model derived from
TanDEM-X and independent of the LiDAR used by both benchmarks. GLO-30 is a surface model and includes part of the forest
and building height. We remove this part with the predicted objects,
\begin{equation}
  \hat g = D - \beta\, \mathcal{B}_{30}(\hat h), \qquad \hat H = \hat g + \hat h,
  \label{eq:ground}
\end{equation}
where $D$ is GLO-30 on the image grid, $\mathcal{B}_{30}$ is a $31\times31$\,m box filter and $\beta = 1$. Here $\hat g$ estimates the bare ground, from which slopes are computed, and $\hat H$ the absolute surface.

\subsection{GeoDSL Programs}
\label{sec:geodsl}

\noindent\textbf{Language.}
GeoDSL has eleven layers, each holding one measured quantity, namely \dsl{OBJECT\_HEIGHT} ($\hat h$), \dsl{HEIGHT} ($\hat H$),
\dsl{GROUND} ($\hat g$), \dsl{SLOPE}, \dsl{SVF}, the masks \dsl{BUILDING}, \dsl{TREE} and \dsl{NATURE}, and three image
layers for edges, gradient and brightness. Measurement functions turn a layer and an option into a number, such as a
regional mean, a point value, an object count, the visible fraction of a viewshed or the clearance of a sight line
(Supp.~Tab.~\ref{tab:supp_dsl}). Numbers combine by arithmetic, comparison, \dsl{min}, \dsl{max} and \dsl{ifelse}, and one answer
constructor closes the program. \dsl{pick\_max} and \dsl{pick\_min} choose an option, \dsl{rank\_desc} and
\dsl{rank\_asc} order them, \dsl{value} reports the target's score rounded to a step and unit, and \dsl{yes\_no} tests a
condition. A complete program is one expression. The mean building height in a region, for instance, is
\begin{center}
\dsl{value(masked\_mean(OBJECT\_HEIGHT, BUILDING, o), step=1, unit="m")}
\end{center}
A static checker infers whether every subexpression is a layer, a number, a condition or an option, and rejects a
program
whose arguments have the wrong kind, whose function needs a point but receives a region, or whose constructor does not
match the answer form. The executor evaluates an accepted program for every option and forms the answer, so a program
cannot return a malformed answer.

\noindent\textbf{Writing programs.}
A frozen Qwen3-4B-Instruct-2507~\cite{qwen3} receives the question text and a prompt, never the image, and decodes
greedily. The prompt has three parts. The grammar lists the layers and function signatures with one generic example.
Metric notes define the quantities that the benchmarks name, such as the sky view factor or the
openness of an area, as expressions over the layers. For Geo3DVQA they transcribe the scoring formulas of its paper,
which the image-only baselines do not receive. Execution feedback returns the checker's message or a runtime error
to the writer, which has two rounds to repair its program. A program that still fails yields no answer and is scored wrong.

\subsection{Scale-Invariance Check}
\label{sec:cert}

Write $\hat h = a\,h$ for a tile-wide factor $a>0$, which models the scale ambiguity of a monocular estimator and removes half or more of its squared error on our test tiles (\supref{sec:supp_sensors}), and let every layer derived from $\hat h$ be
recomputed. A program's answer is \emph{scale-invariant} if it is the same for every $a$. We check this statically, much as in dimensional analysis,
by giving every subexpression a degree in $\hat h$. Means, point values, window means, masked means and object maxima
of \dsl{OBJECT\_HEIGHT} have degree~1, as do standard deviations, and variances have degree~2. Masks and image layers
come from other sensors and have degree~0, like constants, and every mask and window radius passed to a function
must have degree~0, so that the sampled pixels do not move with~$a$. \dsl{GROUND} and \dsl{SLOPE} subtract the blurred
$\hat h$ of Eq.~\eqref{eq:ground}, and \dsl{SVF}, \dsl{HEIGHT}, viewsheds and clearances mix $\hat h$ with other
quantities, so all of them have no degree. A sum or difference needs equal degrees, products and quotients add and
subtract them, and \dsl{abs}, \dsl{min} and \dsl{max} keep them. \dsl{ifelse} keeps the degree of its branches only
if its condition compares two terms of equal degree, and \dsl{exp} accepts only degree~0.
\begin{equation}
  \text{INV}(P) \iff
  \begin{cases}
    \deg(s)\ \text{is defined} & \text{for \dsl{pick} and \dsl{rank}},\\
    \deg(s) = 0 & \text{for \dsl{value}},\\
    \deg(\ell) = \deg(r) & \text{for \dsl{yes\_no}}(\ell \lessgtr r),
  \end{cases}
  \label{eq:cert}
\end{equation}
where $s$ is the option score. A score of degree $k$ becomes $a^k s$ under the stretch, which preserves the order of the
options for $a>0$, so every program with $\text{INV}(P)$ returns the same answer for any positive rescaling of
$\hat h$. The check concerns the height scale alone, so segmentation errors and height errors that vary within a tile reach invariant answers as they reach the others (Sec.~\ref{sec:analysis}).

\section{\bench{}}
\label{sec:bench}

\noindent\textbf{Sources and tiles.}
\bench{} pairs orthophotos with airborne LiDAR surface and terrain models and with a national topographic vector map
in two countries. AltiBench-CH uses SWISSIMAGE~\cite{swissimage}, swissSURFACE3D~\cite{swisssurface}, swissALTI3D~\cite{swissalti} and swissTLM3D~\cite{swisstlm} from swisstopo, and AltiBench-FR uses BD ORTHO~\cite{bdortho}, the LiDAR HD surface and terrain models~\cite{lidarhd} and BD TOPO~\cite{bdtopo} from IGN. Each country contributes six areas, Zurich, Geneva, Basel, Bern, Lugano and Davos in
Switzerland and Grenoble, Annecy, Chamb\'ery, Chamonix, Brian\c{c}on and Nice in France, from dense city centres to
Alpine valleys. We resample all layers to 1\,km tiles at 1\,m and draw ten test and four development tiles per area
with a fixed seed, keeping development tiles away from test tiles (\supref{sec:supp_bench}).

\noindent\textbf{Questions and answers.}
\bench{} has 15 question types (Supp.~Tab.~\ref{tab:types}). They ask for building and tree heights, building counts,
cover fractions and volumes, and the sky view factor at a point or over regions. Others ask for the best rooftop for
solar panels, the lowest area, the order of areas by slope, the mean ground elevation of a region and whether two
points see each other.
A stand-alone script computes every answer from the LiDAR models and the mapped building footprints. Choices require a margin between the best and the second-best option, such as 5\,m of ground
elevation, and numbers count as correct within a tolerance, such as $\max(2\,\text{m}, 20\%)$ for heights. Each type
has two to five phrasings, written by OLMo-2-7B~\cite{olmo2}, a model family outside the systems we test, and checked
by hand, and a definition line states how the quantity is computed. We checked at least 50 random questions per country against the reference layers (\supref{sec:supp_bench}).

\begin{table}[t]
\centering
\footnotesize
\setlength{\tabcolsep}{1.1pt}
\begin{tabular}{@{}lllrccc@{}}
\toprule
Benchmark & View & Regions & Test & Metric & Terrain & Script \\
\midrule
RSVQA-HR~\cite{rsvqa} & aerial & US & -- & & & \checkmark \\
SpatialRGPT-Bench~\cite{spatialrgpt} & ground & -- & 1,406 & \checkmark & & \\
SpatialSky-Bench~\cite{spatialsky} & UAV & -- & 1,000 & part & & \\
VertiCue-Bench~\cite{verticue} & aerial & US & 1,534 & part & & \checkmark \\
GeoHeight-Bench/+~\cite{geoheight} & aerial & DE/FR/US & -- & \checkmark & \checkmark & \\
Geo3DVQA~\cite{geo3dvqa} & aerial & DE & 10,232 & \checkmark & \checkmark & part \\
\midrule
\textbf{\bench{}} & aerial & CH/FR & 2,889 & \checkmark & \checkmark & \checkmark \\
\bottomrule
\end{tabular}
\caption{\textbf{Related benchmarks.} \emph{Metric} marks metric 3D answers, \emph{Terrain} questions on ground elevation or slope, and \emph{Script} answers computed from reference data and scored without an LLM. Dashes mark values not reported.}
\label{tab:benchcmp}
\end{table}

\noindent\textbf{Comparison.}
Tab.~\ref{tab:benchcmp} places \bench{} among related benchmarks. GeoHeight-Bench/+ and the free-form part of Geo3DVQA include answers written or judged by language
models. \bench{} is the only one in the table whose answers all come from LiDAR and national maps by a script, across two
countries, with metric numbers, terrain and line-of-sight questions.

\noindent\textbf{Splits and use.}
The test sets hold 1,483 Swiss and 1,406 French questions. Half of the answers are numbers, and the rest are
choices, orderings or yes and no. Always giving the most frequent answer of each type scores 33.0\%, and a VLM that
reads only the question text scores below it. The generator also produces 3,357 Swiss training questions on further
tiles, which we use for an in-domain fine-tuned baseline. LiDAR serves only to construct reference answers, and
\method{} needs the orthophoto and a global elevation model, so it applies where LiDAR is unavailable.

\begin{table*}[t]
\centering
\small
\setlength{\tabcolsep}{3.6pt}
\begin{tabular}{@{}lcccll@{}}
\toprule
Method & Fine-tuned & AltiBench-CH & AltiBench-FR & AltiBench & \method{} $-$ method \\
\midrule
Constant answer per type & -- & 30.7 & 35.3 & 33.0 & \pp{+39.6} \ci{37.4}{41.7} \\
\midrule
\multicolumn{6}{@{}l}{\textit{Fine-tuned on 3D or spatial question data}} \\
\trained Qwen2.5-VL-7B fine-tuned on Geo3DVQA~\cite{geo3dvqa} & \checkmark & \underline{27.5} & \underline{28.1} & \underline{27.8} & \pp{+44.8} \ci{42.6}{46.9} \\
\trained Qwen2.5-VL-7B fine-tuned on \bench{}-CH & \checkmark & \textbf{41.7} & \textbf{41.0} & \textbf{41.3} & \pp{+31.2} \ci{28.6}{33.8} \\
\trained SpaceThinker~\cite{spacethinker} & \checkmark & 21.0 & 24.2 & 22.6 & \pp{+50.0} \ci{47.9}{52.0} \\
\midrule
\multicolumn{6}{@{}l}{\textit{Remote sensing VLMs, image only}} \\
\tfree EarthDial~\cite{earthdial} & $\times$ & \textbf{20.2} & \textbf{17.7} & \textbf{19.0} & \pp{+53.5} \ci{51.4}{55.6} \\
\tfree GeoChat~\cite{geochat} & $\times$ & \underline{12.1} & \underline{13.0} & \underline{12.6} & \pp{+60.0} \ci{58.2}{61.8} \\
\midrule
\multicolumn{6}{@{}l}{\textit{Open-weight VLMs, image only}} \\
\tfree Qwen2.5-VL-7B~\cite{qwen25vl} & $\times$ & 26.8 & 29.0 & 27.9 & \pp{+44.7} \ci{42.5}{46.8} \\
\tfree Qwen3-VL-8B~\cite{qwen3vl} & $\times$ & 31.4 & 28.9 & 30.2 & \pp{+42.4} \ci{40.2}{44.5} \\
\tfree Qwen3-VL-32B~\cite{qwen3vl} & $\times$ & 39.0 & 36.2 & 37.7 & \pp{+34.9} \ci{32.8}{37.0} \\
\tfree Qwen3.6-27B~\cite{qwen36} & $\times$ & \underline{42.9} & 40.8 & \underline{41.8} & \pp{+30.7} \ci{28.1}{33.3} \\
\tfree Qwen3.6-35B-A3B~\cite{qwen36moe} & $\times$ & 42.8 & 39.7 & 41.3 & \pp{+31.3} \ci{29.1}{33.5} \\
\tfree GLM-4.5V~\cite{glm45v} & $\times$ & 38.2 & 36.5 & 37.4 & \pp{+35.2} \ci{32.8}{37.5} \\
\tfree GLM-5.3-Flash~\cite{glm53} & $\times$ & 41.3 & \underline{42.0} & 41.6 & \pp{+30.9} \ci{28.5}{33.3} \\
\tfree Gemma 4 26B-A4B~\cite{gemma4} & $\times$ & 42.4 & 39.5 & 41.0 & \pp{+31.6} \ci{29.2}{33.9} \\
\tfree Gemma 4 31B~\cite{gemma4} & $\times$ & \textbf{44.9} & \textbf{45.2} & \textbf{45.1} & \pp{+27.5} \ci{25.0}{29.9} \\
\tfree Nemotron 3.5 Super VL~\cite{nemotron} & $\times$ & 39.9 & 40.1 & 40.0 & \pp{+32.5} \ci{30.4}{34.8} \\
\midrule
\multicolumn{6}{@{}l}{\textit{Proprietary VLMs, image only}} \\
\tfree Qwen3.7-Plus~\cite{qwen37} & $\times$ & 49.0 & 46.6 & 47.8 & \pp{+24.7} \ci{22.4}{27.2} \\
\tfree Gemini 3.8 Flash~\cite{gemini38} & $\times$ & \textbf{62.6} & \textbf{61.5} & \textbf{62.1} & \pp{+10.5} \ci{8.3}{12.7} \\
\tfree GPT-6.1 Terra~\cite{gpt61} & $\times$ & \underline{60.3} & \underline{60.1} & \underline{60.2} & \pp{+12.4} \ci{10.2}{14.5} \\
\midrule
\multicolumn{6}{@{}l}{\textit{Programs over our sensors}} \\
\tfree VisProg-style~\cite{visprog}, our sensors & $\times$ & \textbf{58.7} & \textbf{60.0} & \textbf{59.4} & \pp{+13.2} \ci{12.1}{14.3} \\
\tfree ViperGPT-style~\cite{vipergpt}, our sensors & $\times$ & \underline{55.8} & \underline{56.6} & \underline{56.2} & \pp{+16.4} \ci{15.0}{17.8} \\
\midrule
\textbf{\method{} (ours)} & $\times$ & \textbf{72.4} & \textbf{72.7} & \textbf{72.6} \ci{71.0}{74.1} & \\
\midrule
\multicolumn{6}{@{}l}{\textit{Reading \method{}'s measurements}} \\
\tfree Qwen3-4B + measurements, text only & $\times$ & 61.0 & 64.4 & 62.7 & \pp{+9.9} \ci{8.6}{11.1} \\
\tfree Qwen2.5-VL-7B + measurements & $\times$ & \underline{66.1} & \underline{66.6} & \underline{66.4} & \pp{+6.2} \ci{4.9}{7.5} \\
\tfree Qwen3-VL-8B + measurements & $\times$ & 55.2 & 59.5 & 57.3 & \pp{+15.3} \ci{13.7}{16.9} \\
\tfree GPT-6.1 Terra + heights only & $\times$ & 61.6 & 65.2 & 63.3 & \pp{+9.2} \ci{7.2}{11.2} \\
\tfree GPT-6.1 Terra + height maps as images & $\times$ & 61.2 & 65.9 & 63.4 & \pp{+9.1} \ci{7.0}{11.1} \\
\tfree GPT-6.1 Terra + measurements & $\times$ & \best{\textbf{72.8}} & \best{\textbf{75.7}} & \best{\textbf{74.2}} & \pp{-1.7} \ci{-2.5}{-0.8} \\
\method{} on reference LiDAR maps & $\times$ & 90.2 & 91.4 & 90.8 & \\
\bottomrule
\end{tabular}
\caption{\textbf{Accuracy (\%) on \bench{}.} Beige rows are fine-tuned on 3D or spatial question data, green rows are training-free baselines. \textbf{Bold} and
\underline{underline} mark the best two per group, \hlword{yellow} the best overall, excluding reference maps. The
last column is the paired difference on the pooled set.}
\label{tab:main}
\end{table*}

\begin{figure*}[t]
\centering
\includegraphics[width=\textwidth]{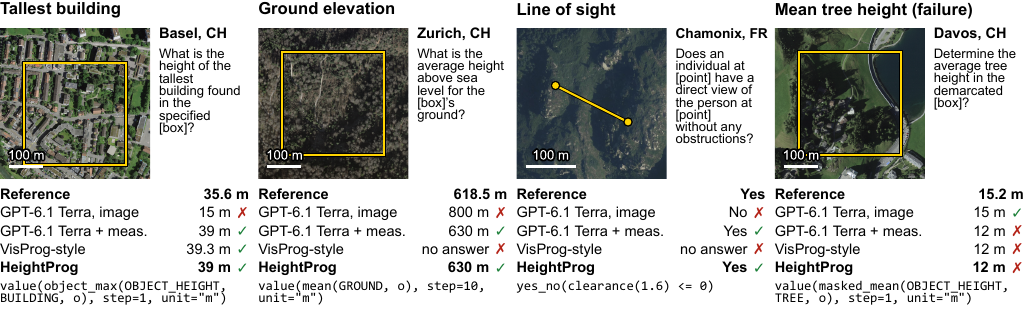}
\caption{\textbf{Answers on four \bench{} test questions.} \method{} answers the left three correctly and GPT-6.1 Terra
on the image does not. On the right, the program is the defined one, but the height sensor underestimates the trees.
$\checkmark$ marks answers within the scoring tolerance. Coordinates in the questions are shown as [box] and [point]
and drawn in yellow. Random examples of every type are in Supp.~Figs.~\ref{fig:gallery1} to~\ref{fig:gallery3}.}
\label{fig:compare}
\end{figure*}

\section{Experiments}
\label{sec:exp}

\subsection{Setup}
\label{sec:setup}

\noindent\textbf{Benchmarks.}
We evaluate on the two \bench{} test sets and report them separately and pooled. On
Geo3DVQA~\cite{geo3dvqa} we use the full short-answer test set of 10,232 questions on 333 GeoNRW images and its
official scorer. For street-level images we use SpatialRGPT-Bench~\cite{spatialrgpt}, split by frame into
development and test frames, with Depth Pro~\cite{depthpro} as the sensor (Sec.~\ref{sec:street}). We chose every rule, prompt and setting on development splits that share no
tile or frame with the test sets.

\noindent\textbf{Baselines.}
Image-only VLMs, from open-weight models to GPT-6.1 Terra and Gemini 3.8 Flash, receive the image and the question
with an instruction to give a short answer. Remote sensing and spatial specialists are EarthDial~\cite{earthdial}, GeoChat~\cite{geochat}
and SpaceThinker~\cite{spacethinker}. Trained baselines fine-tune Qwen2.5-VL-7B with LoRA~\cite{lora} following the
recipe of Geo3DVQA, once on its 100K training questions and once on the 3,357 \bench{} training questions. A
VLM given \method{}'s measurements receives, as text next to the image, every quantity that our functions compute for the
question's options from the same layers and ground. In further variants, GPT-6.1 Terra reads only the heights and ground elevations or also sees these layers as images, and our writer Qwen3-4B reads the text without the image. VisProg-style~\cite{visprog} and ViperGPT-style~\cite{vipergpt}
programs use the same writer, metric notes, repair rounds, layers and functions, but write a step list or free Python instead of one typed
expression. We normalise baseline answer tags, special tokens and option letters before the shared scorer (\supref{sec:supp_baselines}) and switch reasoning off wherever a model allows it.

\noindent\textbf{Metrics.}
Accuracy counts a numeric answer as correct within the tolerance of its type. Intervals are 95\% bootstrap intervals
with 10,000 resamples over tiles, or over images on Geo3DVQA and SpatialRGPT-Bench, and differences between two
systems are paired over the same questions.

\subsection{Main Results}
\label{sec:main}

\noindent\textbf{\bench{}.}
\method{} outperforms every image-only and fine-tuned model in both countries and answers 72.6\% of the pooled
\bench{} questions (Tab.~\ref{tab:main}, Fig.~\ref{fig:compare}). Of the lead over GPT-6.1 Terra, the three ground types, whose answers no image shows, give 5.7 points, and mean tree height costs 3.0 (Fig.~\ref{fig:types}). The Geo3DVQA fine-tuned model and the remote sensing and spatial specialists score below the constant, and the
same recipe trained on \bench{}-CH stays below the frontier models that see only the image. The ranks of the systems agree across the two countries with a Spearman
correlation of 0.97.

\begin{table}[t]
\centering
\small
\setlength{\tabcolsep}{2.5pt}
\begin{tabular}{@{}lcrl@{}}
\toprule
Method & FT & \multicolumn{2}{c}{Accuracy} \\
\midrule
Constant answer per category & -- & 29.3 & \\
\midrule
\trained Qwen2.5-VL-7B fine-tuned, published~\cite{geo3dvqa} & \checkmark & \underline{49.6} & \\
\trained Qwen2.5-VL-7B fine-tuned, reproduced & \checkmark & \textbf{55.6} & \\
\midrule
\tfree GPT-4o, published~\cite{geo3dvqa} & $\times$ & 28.6 & \\
\tfree Gemini-2.5-Flash, published~\cite{geo3dvqa} & $\times$ & 33.0 & \\
\tfree Qwen2.5-VL-7B~\cite{qwen25vl} & $\times$ & 25.1 & \\
\tfree Qwen3-VL-8B~\cite{qwen3vl} & $\times$ & 25.2 & \\
\tfree Qwen3.6-35B-A3B~\cite{qwen36moe} & $\times$ & 41.0 & \\
\tfree Qwen3.6-27B~\cite{qwen36} & $\times$ & \underline{41.3} & \\
\tfree GPT-6.1 Terra~\cite{gpt61} & $\times$ & \textbf{56.6} & \\
\midrule
\textbf{\method{} (ours)} & $\times$ & \best{\textbf{74.6}} & \ci{73.3}{75.8} \\
\bottomrule
\end{tabular}
\caption{\textbf{Accuracy (\%) on the Geo3DVQA test set} with the
official scorer. FT, fine-tuned on 3D or spatial question data. Marking as in Tab.~\ref{tab:main}.}
\label{tab:g3d}
\end{table}

\noindent\textbf{Geo3DVQA.}
On Geo3DVQA, \method{} reaches 74.6\% (Tab.~\ref{tab:g3d}), 25 points above the published fine-tuned model, and GPT-6.1
Terra with the image alone exceeds both this model and our stronger reproduction. The lead comes from the ten categories that a program computes, where \method{} answers 79.4\%
against 51.4\% for the reproduced model, and leads with the grammar alone (Tab.~\ref{tab:ablation},
\supref{sec:supp_types}).

\subsection{Giving VLMs the Measurements}
\label{sec:matched}

\begin{figure}[t]
\centering
\includegraphics[width=\linewidth]{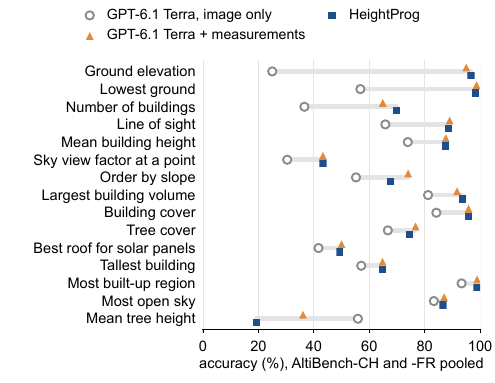}
\caption{\textbf{Accuracy per question type.} Measurements help GPT-6.1 Terra most on ground elevations and counts.
\method{} follows this pattern except for mean tree height. Markers are offset vertically.}
\label{fig:types}
\end{figure}

In the lower block of Tab.~\ref{tab:main}, VLMs receive the quantities that our functions compute, and every one of
them gains. GPT-6.1 Terra rises from 60.2\% to 74.2\% and Qwen2.5-VL-7B more than doubles (Fig.~\ref{fig:types}).
From the image alone, the two frontier models already answer 84 to 93\% of the building-cover questions, so they
locate the regions. Heights and ground elevations alone, as text or as images, settle ground elevation, while counts,
slope order and line of sight need the derived quantities. Our writer Qwen3-4B answers better by writing programs
than by reading the measurements as text, most on building counts and mean building height. \method{} is within one point of the
best reader of our sensors on 11 of the 15 types (\supref{sec:supp_types}). GPT-6.1 Terra reading the measurements is
1.7 points ahead pooled and level in Switzerland, and mean tree height and slope order account for this lead. For tree
height, nearly half of our programs compute another quantity and the height sensor underestimates tall trees, which the
VLM corrects with its prior. For slope order, the 30\,m elevation model is too coarse.

\subsection{Street-Level Images}
\label{sec:street}

\begin{table}[t]
\centering
\small
\setlength{\tabcolsep}{3.5pt}
\begin{tabular}{@{}lcccc@{}}
\toprule
Method & FT & Qual. & Quant. & All \\
\midrule
\trained SpatialRGPT-VILA-1.5-8B$^\dagger$~\cite{spatialrgpt} & \checkmark & \underline{92.7} & \underline{62.7} & -- \\
\trained SpatialReasoner-R1 8B$^\dagger$~\cite{spatialreasoner} & \checkmark & \best{\textbf{95.6}} & \best{\textbf{77.3}} & -- \\
\midrule
\tfree Qwen2.5-VL-7B~\cite{qwen25vl} & $\times$ & 72.2 & 29.5 & 49.9 \\
\tfree Qwen3-VL-8B~\cite{qwen3vl} & $\times$ & 84.8 & 44.4 & 63.7 \\
\tfree Qwen3.6-27B~\cite{qwen36} & $\times$ & \textbf{90.9} & \underline{46.1} & \underline{67.5} \\
\tfree GPT-6.1 Terra~\cite{gpt61} & $\times$ & \textbf{90.9} & \textbf{47.0} & \textbf{68.0} \\
\midrule
\multicolumn{5}{@{}l}{\textit{Programs over our sensor}} \\
\tfree ViperGPT-style~\cite{vipergpt}, our sensor & $\times$ & 88.9 & 72.7 & 80.5 \\
\midrule
\textbf{\method{} (ours)} & $\times$ & \textbf{92.6} & \textbf{72.4} & \textbf{82.0} \\
\midrule
\multicolumn{5}{@{}l}{\textit{Reading \method{}'s measurements}} \\
\tfree Qwen3-VL-8B + measurements & $\times$ & \underline{87.0} & \underline{70.3} & \underline{78.3} \\
\tfree GPT-6.1 Terra + measurements & $\times$ & \textbf{93.6} & \textbf{71.8} & \best{\textbf{82.2}} \\
\bottomrule
\end{tabular}
\caption{\textbf{Accuracy (\%) on SpatialRGPT-Bench,} our test split of 1,018 questions. $^\dagger$Published numbers on
all 1,406 questions with a GPT-4 judge. FT, fine-tuned on 3D or spatial question data. Marking as in Tab.~\ref{tab:main}.}
\label{tab:street}
\end{table}

On SpatialRGPT-Bench, \method{} answers 82.0\% of the test questions (Tab.~\ref{tab:street}), 14 points
above GPT-6.1 Terra with the image alone, level with the same model reading our measurements and 1.5 points above
ViperGPT-style free Python over the same sensor. Depth Pro replaces the height estimator, and ten scene functions for distances, sizes
and directions extend the typed language (\supref{sec:supp_street}). Models trained on spatial question data remain ahead on qualitative comparisons and on object widths
and heights, while \method{} leads on all three distance types.

\subsection{Error Analysis}
\label{sec:analysis}

\begin{table}[t]
\centering
\small
\setlength{\tabcolsep}{3.5pt}
\begin{tabular}{@{}lccc@{}}
\toprule
Inputs to the same programs & CH & FR & Geo3DVQA$^*$ \\
\midrule
All predicted (\method{}) & 72.4 & 72.7 & 79.4 \\
Reference land cover & 75.2 & 77.0 & 80.0 \\
Reference heights, sky view, ground & 85.5 & 85.8 & 93.6 \\
All reference & 90.2 & 91.4 & 94.1 \\
\midrule
Share of the gap closed by geometry & 74\% & 70\% & 97\% \\
\bottomrule
\end{tabular}
\caption{\textbf{Sensor error decomposition.} We execute the programs of \method{} with one group of predicted
layers replaced by the reference. $^*$The ten categories that a program computes.}
\label{tab:decomp}
\end{table}

\noindent\textbf{Error decomposition.}
The geometry sensors limit \method{} more than the segmentation does. Running the same programs on reference heights,
sky view factors and ground closes 74\% and 70\% of the gap between predicted and reference inputs in the two countries
and 97\% on Geo3DVQA (Tab.~\ref{tab:decomp}). Reference land cover closes much less. Mean tree height is the exception, where wrong programs cost as much as the sensor (\supref{sec:supp_types}).

\begin{figure}[t]
\centering
\includegraphics[width=\linewidth]{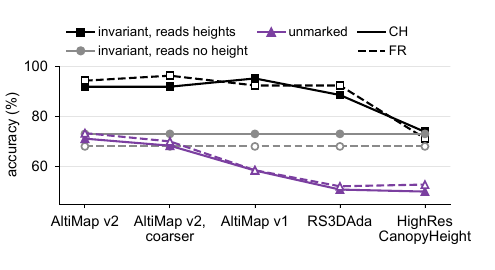}
\caption{\textbf{The same programs under five height estimators,} ordered by overall accuracy. The invariant answers that read
heights are 113 volume comparisons. The 747 invariant answers that read no height keep their accuracy under every estimator.}
\label{fig:sensors}
\end{figure}

\noindent\textbf{Scale invariance.}
On the Swiss development set, rescaling each tile's predicted heights by a random factor between 0.5 and 2
changes none of the answers that the check marks. On the test sets it marks 30\% of the answers. Among them, the volume comparisons
read heights, and across four estimators their accuracy varies by at most 7 points, against 20 to 21 for unmarked
answers. The others read no height, such as covers and counts, or are 98 tree-height programs that answer in metres
from the tree mask, a dimensional error that the degrees expose. A canopy height model~\cite{hrch} lowers both
groups (Fig.~\ref{fig:sensors}). At street level,
every marked answer reads the depth (\supref{sec:supp_street}). \bench{} also ranks height estimators by their use in reasoning. The five keep the same order in both
countries (Fig.~\ref{fig:sensors}), and agreement with the answers on reference maps, which needs no answer labels,
reproduces it up to one swap of two nearly tied estimators.

\subsection{Ablation Studies}
\label{sec:ablation}

\begin{table}[t]
\centering
\small
\setlength{\tabcolsep}{3.5pt}
\begin{tabular}{@{}lccc@{}}
\toprule
Variant & CH & FR & Geo3DVQA$^*$ \\
\midrule
\method{}, Qwen3-4B writer & \textbf{72.4} & \textbf{72.7} & \textbf{79.4} \\
\midrule
\multicolumn{4}{@{}l}{\textit{Prompt}} \\
\quad grammar only & 48.2 & 49.6 & 59.1 \\
\quad + metric notes, no feedback & \underline{66.1} & \underline{67.4} & \underline{79.3} \\
\multicolumn{4}{@{}l}{\textit{Program form, same writer and sensors}} \\
\quad VisProg-style step list & 58.7 & 60.0 & 43.1 \\
\quad ViperGPT-style free Python & 55.8 & 56.6 & 76.7 \\
\multicolumn{4}{@{}l}{\textit{Sensors}} \\
\quad no ground sensor & 60.9 & 61.1 & 72.5 \\
\bottomrule
\end{tabular}
\caption{\textbf{Ablations} on the test sets (\%). Each block changes one component. Bold and underline mark the
best two per column. Other writers are in \supref{sec:supp_writers}. $^*$The ten program categories.}
\label{tab:ablation}
\end{table}

\noindent\textbf{Prompt and program form.}
The metric notes are the most important part of the prompt. With only the grammar, the 4B writer produces invalid
programs for a third of the \bench{} questions. The notes recover most of the loss, and the execution feedback adds the rest
on \bench{} and nothing on Geo3DVQA. Given the same
layers and functions, a VisProg-style step list loses 13 points on \bench{} and 36 on Geo3DVQA, and ViperGPT-style free Python loses 16 and 2.7 (Tab.~\ref{tab:ablation}). These programs compute the wrong quantity, mix layers with
numbers or return an answer in the wrong form. Free Python needs a large writer and GeoDSL does not. With
DeepSeek-V4.1-Flash writing both, GeoDSL and free Python are level, and the GeoDSL programs take 22 output tokens
against 59. With our 4B writer, free Python drops by 19 points and GeoDSL by 2 (\supref{sec:supp_writers}).

\noindent\textbf{Ground sensor and writer.}
Without the elevation model, the questions on ground elevation, lowest areas and slope have no ground to read, and
\method{} loses on both benchmarks. Of eight other writers, six score lower than Qwen3-4B, four of them
through invalid programs, and two larger models score higher (\supref{sec:supp_writers}). On
test questions of four new composite types that no prompt or example covers, \method{} answers 83\% correctly and
writes the defined program for 76\%, with tree questions the weakest (\supref{sec:supp_types}).

\begin{table}[t]
\centering
\footnotesize
\setlength{\tabcolsep}{2.2pt}
\begin{tabular}{@{}lrrrrr@{}}
\toprule
& \multicolumn{3}{c}{Tokens} & Output & \\
\cmidrule(lr){2-4}
API model, image & Input & Output & Reasoning & per question & Acc. \\
\midrule
Qwen3-VL-32B & 3.50M & 1.45M & 0 & 501 & 37.7 \\
Qwen3.6-27B & 3.54M & 1.63M & 0 & 564 & 41.8 \\
Qwen3.6-35B-A3B & 3.55M & 2.27M & 0 & 786 & 41.3 \\
GLM-4.5V & 4.44M & 0.02M & 0 & 8 & 37.4 \\
GLM-5.3-Flash & 4.45M & 0.38M & 0.01M & 131 & 41.6 \\
Gemma 4 26B-A4B & 1.52M & 1.13M & 0 & 390 & 41.0 \\
Gemma 4 31B & 1.51M & 0.96M & 0 & 333 & 45.1 \\
Nemotron 3.5 Super VL & 4.15M & 2.45M & 0 & 848 & 40.0 \\
Qwen3.7-Plus & 3.54M & 2.83M & 0 & 978 & 47.8 \\
Gemini 3.8 Flash & 3.91M & 3.86M & 3.56M & 1,335 & 62.1 \\
GPT-6.1 Terra & 4.23M & 0.05M & 0 & 16 & 60.2 \\
\quad + heights only & 4.54M & 0.05M & 0 & 16 & 63.3 \\
\quad + height maps & 12.43M & 0.05M & 0 & 16 & 63.4 \\
\quad + measurements & 4.78M & 0.05M & 0 & 16 & 74.2 \\
\bottomrule
\end{tabular}
\caption{\textbf{Tokens of the API runs} on both \bench{} test sets, with the pooled accuracy (\%). The last three rows
add our heights, height maps or all measurements to the image.}
\label{tab:supp_tokens}
\end{table}

\noindent\textbf{Cost.}
On an RTX 3090, the sensors take 23\,s per tile, and writing and running a program 0.19\,s per question. \bench{}-CH takes 28 minutes, against 49 for Qwen2.5-VL-7B reading our
measurements (\supref{sec:supp_cost}). Table~\ref{tab:supp_tokens} lists the tokens of every API run. Output tokens per question differ by two
orders of magnitude, and the two frontier models, at 16 and 1,335, differ by 1.9 points in accuracy. Image tokens are counted by each provider and are not comparable across providers.

\section{Conclusion}
\label{sec:conclusion}

VLMs answering metric 3D questions about aerial images are held back by missing geometric measurements. \method{} hands the measuring to frozen
sensors, a monocular height estimator and a public elevation model, and lets a 4B model reason in typed programs
over their layers. Without training, it outperforms every image-only and fine-tuned model we evaluate on \bench{} and Geo3DVQA,
carries over to street-level images with a depth sensor, and marks which answers do not depend on the height scale. A frontier VLM that reads the same measurements is 1.7 points ahead on \bench{}. \bench{} provides LiDAR-derived answers in two countries.

\noindent\textbf{Limitations.}
Sensor errors and wrong programs limit \method{}, which reaches 90.8\% on reference maps, and make mean tree height its
hardest type. The 30\,m elevation model is too coarse for slope order, and semantic land use, which no function
measures, falls back to a prior. \bench{} defines each quantity in the question and leaves open-ended questions out
of scope, and its two European countries leave tropical and arid regions and informal settlements untested.

\section*{LLM Usage Statement}
Language models are part of the method and of the evaluation. \method{} writes its programs with Qwen3-4B, and the
baselines are the models listed in Sec.~\ref{sec:setup}. For \bench{}, OLMo-2-7B suggested candidate question types and
drafted their phrasings. We selected and defined the final types and checked every phrasing by hand.
Outside the method, we used Claude Opus 5.5 and GPT-6.1 Sol for code, experiments, analysis, figures, literature search, drafting and editing. The authors
checked all code, numbers, citations and text and take full responsibility for the content.

{
    \small
    \bibliographystyle{ieeenat_fullname}
    \bibliography{refs}
}

\clearpage
\appendix
\onecolumn
\begin{center}
{\Large\bf HeightProg: Typed Programs over Frozen Sensors\\for Metric 3D Question Answering on Aerial Images\par}
\vspace{0.6em}
{\large Supplementary Material\par}
\end{center}
\vspace{0.4em}

\newcommand{\tocline}[2]{\ref{#1} & #2 \dotfill & \pageref{#1} \\}
\begin{center}
\begin{tabular}{@{}l p{0.62\linewidth} r@{}}
\multicolumn{3}{@{}l}{\textbf{Contents}} \\[2pt]
\tocline{sec:supp_bench}{\bench{} construction details and examples}
\tocline{sec:supp_dsl}{GeoDSL reference and prompt}
\tocline{sec:supp_baselines}{Baseline details}
\tocline{sec:supp_street}{Street-level transfer}
\tocline{sec:supp_writers}{Writer ablation}
\tocline{sec:supp_types}{Per-type results}
\tocline{sec:supp_sensors}{Sensor testbed and scale check}
\tocline{sec:supp_cost}{Run time}
\end{tabular}
\end{center}
\vspace{0.6em}

\section{\bench{} Construction Details}
\label{sec:supp_bench}

\noindent\textbf{Grid and reference layers.}
Every tile covers 1\,km$\times$1\,km at 1\,m per pixel. A tile is eligible when the image and the LiDAR were
acquired at most two years apart, less than 1\% of it lacks data and at least 10\% of it is covered by buildings or
trees. Development tiles never touch a test tile or its eight neighbours. The object height is the LiDAR surface model minus the LiDAR
terrain model. Reference sky view factors are computed with UMEP~\cite{svfcast,umep} on the 1\,m surface model with 153 sky
patches. Slopes are central differences of the terrain model after a $5\times5$ mean filter, in degrees.

\noindent\textbf{Buildings and trees.}
A building is a footprint of the topographic map in a roofed class whose median object height is at least 2\,m, so
the LiDAR confirms that it stands. Footprints under construction and footprints that fail this check are marked
uncertain, and a region that contains any uncertain pixel is not used for a building question. Single buildings for
counting are the 4-connected components of the building mask, since stepped buildings and adjacent old-town houses
carry several overlapping footprints. Components below 20\,m$^2$ are dropped. A tree pixel has an object height above
2.5\,m, lies more than 2\,m from every footprint of any class and is not on a road or railway bridge, which is buffered
by half its nominal width plus 3\,m. Tree components below 5\,m$^2$ are dropped. The generator reads class names of
swissTLM3D, so the BD TOPO layers of France are first mapped to their Swiss equivalents and the rules stay unchanged.

\noindent\textbf{Regions, points and sampling.}
Questions follow the Geo3DVQA format of region and point coordinates. Regions are squares with a side of 100 to 300\,m, drawn uniformly from five sizes, with integer percentage
coordinates. The regions of one question do not overlap. Points have one decimal of a percent, which is one pixel.
Regions and points of the sky view factor types keep at least 100\,m from the tile border, because the sky view factor
of a tile does not see obstacles outside it. Regions of the three terrain types contain at most 10\% water. The generator tries two question slots per type and tile, but one for building cover and for the most
built-up region, and leaves a slot empty when no candidate meets its sampling conditions. Ground elevation questions
are drawn on five tiles per area, 30 per country. Correct options are balanced over positions, and line-of-sight
questions are balanced between yes and no. Table~\ref{tab:supp_rules} lists the answer rule, the sampling condition
and the tolerance of every type.

\begin{table}[h]
\centering
\small
\setlength{\tabcolsep}{4pt}
\begin{tabular}{@{}p{3.2cm}p{6.2cm}p{4.6cm}p{2.3cm}@{}}
\toprule
Type & Answer & Sampling condition & Correct if \\
\midrule
Mean building height & mean object height over building pixels & buildings cover $\geq15\%$ & $\pm H$ \\
Tallest building & largest median object height over footprints of $\geq50$\,m$^2$ at least half inside & $\geq3$ such footprints & $\pm H$ \\
Mean tree height & mean object height over tree pixels & trees cover $\geq20\%$ & $\pm H$ \\
Tree cover & share of tree pixels & balanced over four cover bands & $\pm10$ points \\
Building cover & share of building pixels & balanced over four cover bands & $\pm10$ points \\
Most built-up region & region with the largest building cover & lead $\geq10$ points & exact \\
Number of buildings & buildings at least half inside & 3 to 40 buildings; none 30--70\% inside & $\pm\max(1, 20\%)$ \\
Largest building volume & region with the largest sum of object height over building pixels & lead $\geq25\%$ & exact \\
Sky view factor at a point & mean over the $3\times3$ window & std.\ $\leq0.05$ in the $5\times5$ window; balanced over five bands & $\pm0.1$ \\
Most open sky & region with the largest mean sky view factor & lead $\geq0.05$ & exact \\
Best roof for solar panels & building, given by a point on its roof, with the largest mean sky view factor over the roof eroded by one pixel & roofs $\geq100$\,m$^2$; lead $\geq0.05$ & exact \\
Lowest ground & region with the lowest mean terrain height & lead $\geq5$\,m & exact \\
Order by slope & three regions by mean slope, descending & neighbours $\geq3^\circ$ apart & exact order \\
Ground elevation & mean terrain height of a region above sea level & 30 per country & $\pm10\%$ \\
Line of sight & no obstacle above the sight line between eyes 1.6\,m above the ground, surface sampled every 0.5\,m & both points on open ground, 50--300\,m apart; clearance $<-0.5$\,m or $>1$\,m & exact \\
\bottomrule
\end{tabular}
\caption{\textbf{Answer rules of the 15 question types.} All quantities are computed on the reference layers of
Sec.~\ref{sec:supp_bench}; the rounding requested in the question only fixes the answer format.
$H=\max(2\,\text{m}, 20\%)$ of the reference.}
\label{tab:supp_rules}
\end{table}

\begin{table}[h]
\centering
\small
\setlength{\tabcolsep}{3pt}
\begin{tabular}{@{}llrr@{}}
\toprule
Question type & Answer & CH & FR \\
\midrule
Mean building height & number (m) & 93 & 68 \\
Tallest building & number (m) & 110 & 100 \\
Mean tree height & number (m) & 118 & 120 \\
Tree cover & number (\%) & 120 & 120 \\
Building cover & number (\%) & 60 & 60 \\
Most built-up region & choice of 4 & 52 & 37 \\
Number of buildings & number & 108 & 97 \\
Largest building volume & choice of 4 & 118 & 100 \\
Sky view factor at a point & number & 120 & 120 \\
Most open sky & choice of 4 & 119 & 120 \\
Best roof for solar panels & choice of 4 & 96 & 84 \\
Lowest ground & choice of 4 & 115 & 114 \\
Order by slope & ordering of 3 & 105 & 118 \\
Ground elevation & number (m) & 30 & 30 \\
Line of sight & yes or no & 119 & 118 \\
\midrule
Total (60 tiles per country) & & 1,483 & 1,406 \\
\bottomrule
\end{tabular}
\caption{\textbf{Question types of the \bench{} test sets.} Answers are computed from airborne LiDAR and topographic
maps; numbers are correct within a type-specific tolerance.}
\label{tab:types}
\end{table}

\noindent\textbf{Phrasings and checks.}
OLMo-2-7B~\cite{olmo2} wrote two to five phrasings per type, which we checked by hand for correct meaning. Each question
ends with a line on the image scale and a definition line that states how the quantity is computed. For each bank
we drew at least 50 random questions across the types, drew their regions or points and the reference layers on the image, and
recomputed the non-sky-view answers independently. The checks found no error in the answer code.

\noindent\textbf{Splits and release.}
Besides the test sets of Table~\ref{tab:types}, the development sets hold 596 and 587 questions on 24 tiles per
country, and the Swiss training set 3,357 questions on 135 further tiles of the same areas. Training tiles avoid every test and
development tile and their eight neighbours. We release the question banks, the tile lists with acquisition dates,
the generator and the scorer. The reference layers are open data of swisstopo and IGN and can be rebuilt from the
tile lists.

\begin{figure}[p]
\centering
\includegraphics[width=\linewidth]{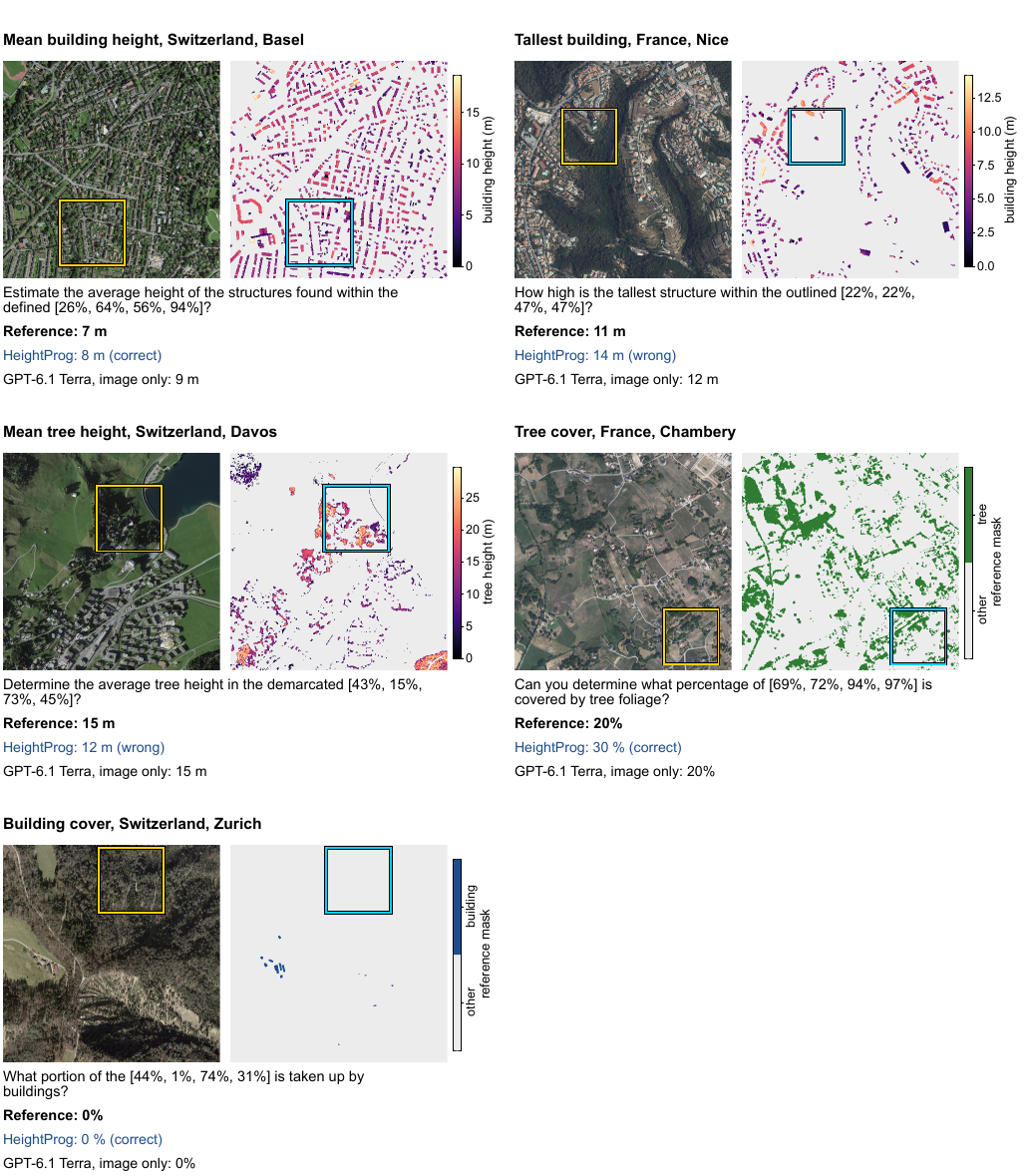}
\caption{\textbf{\bench{} examples, types 1 to 5.} Left, the image with the question geometry. Right, the reference
layer. Questions are cut to their first sentence.}
\label{fig:gallery1}
\end{figure}

\begin{figure}[p]
\centering
\includegraphics[width=\linewidth]{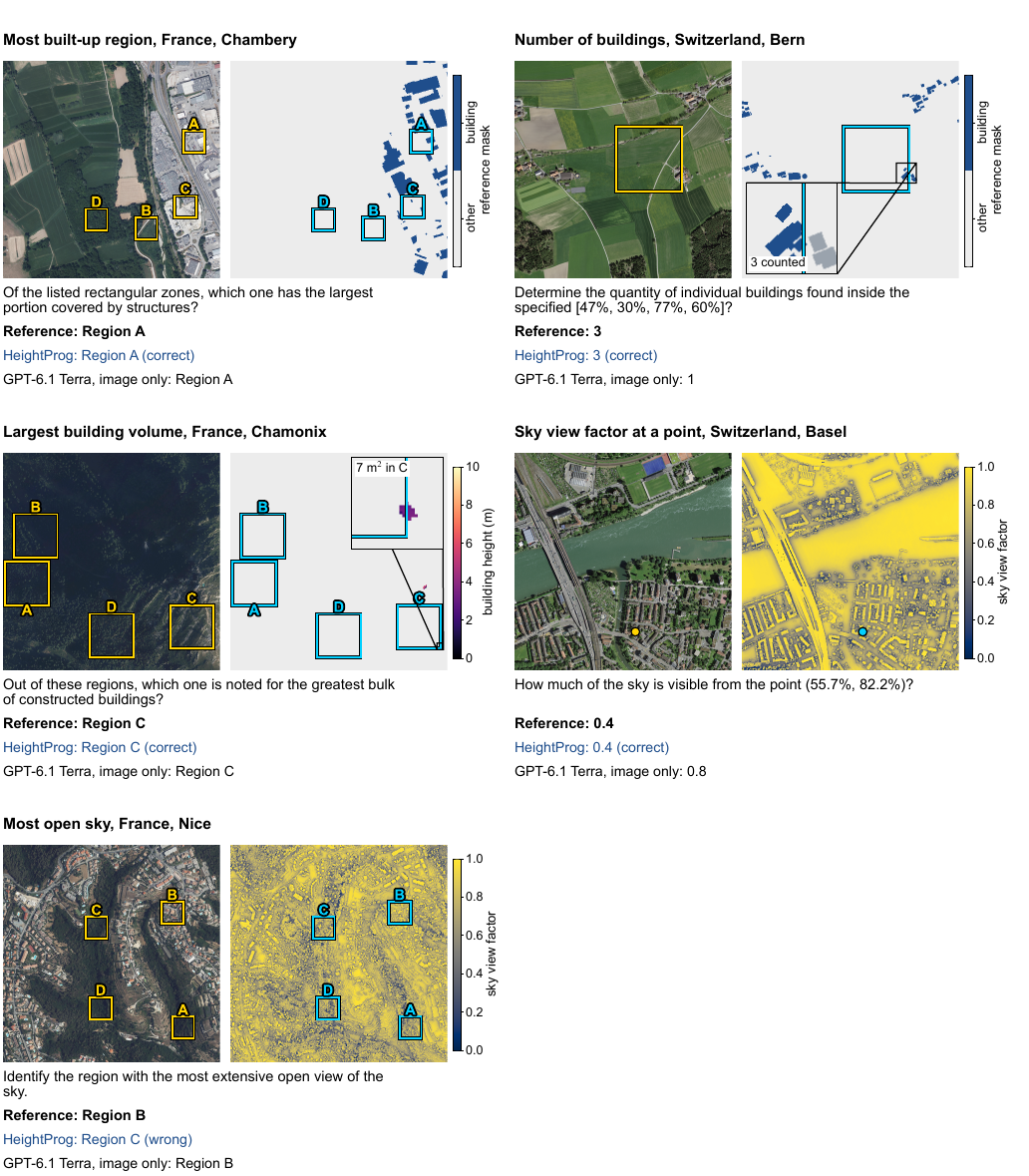}
\caption{\textbf{\bench{} examples, types 6 to 10,} as in Fig.~\ref{fig:gallery1}. Insets magnify the buildings that
decide the count and the volume.}
\label{fig:gallery2}
\end{figure}

\begin{figure}[p]
\centering
\includegraphics[width=\linewidth]{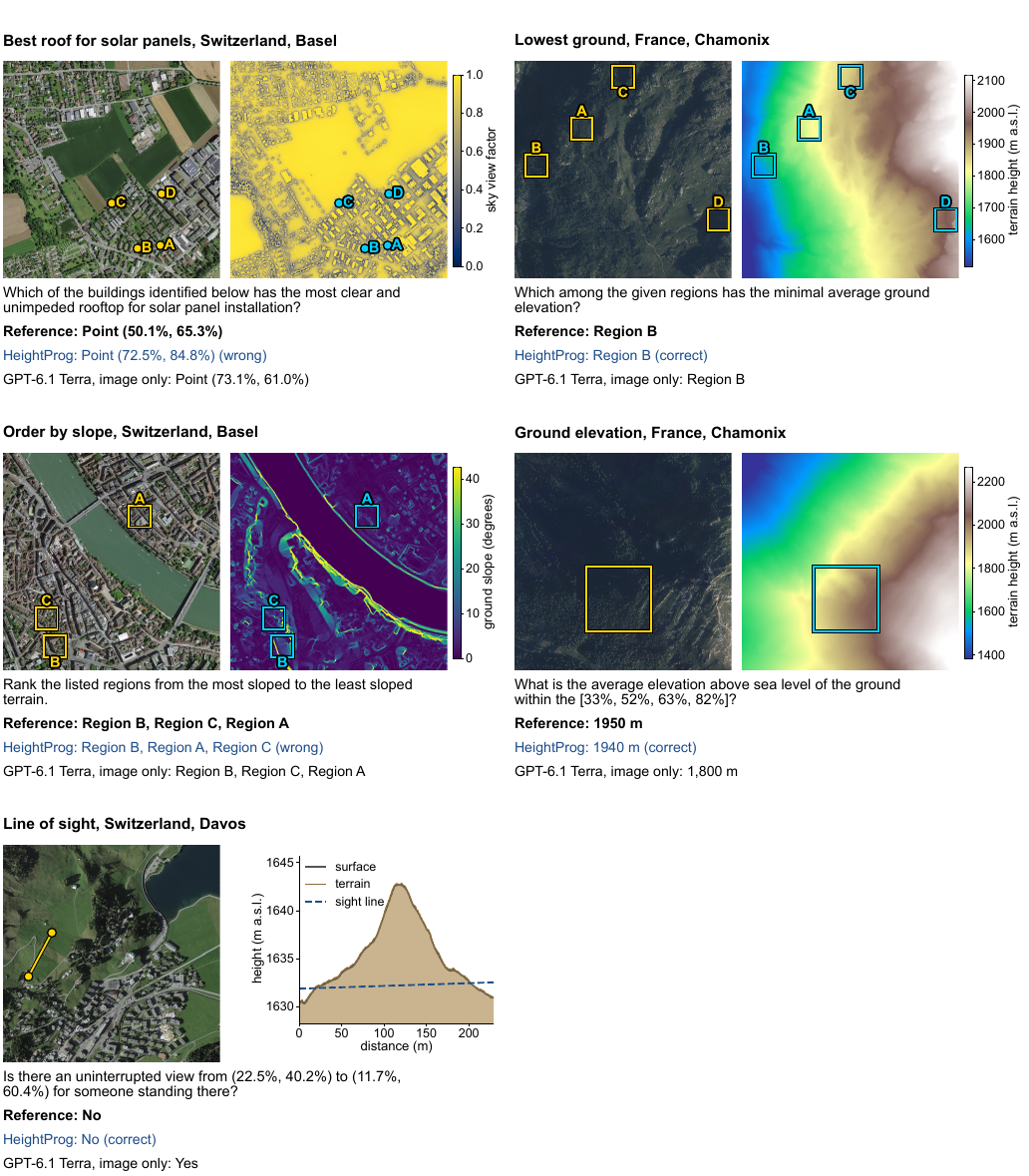}
\caption{\textbf{\bench{} examples, types 11 to 15,} as in Fig.~\ref{fig:gallery1}.}
\label{fig:gallery3}
\end{figure}

\noindent\textbf{Examples.}
Figures~\ref{fig:gallery1} to~\ref{fig:gallery3} show one test question per type, drawn at random with seed 0 and
alternating between the two countries, with the reference layer that its answer is computed from. For covers, counts and the most built-up region the layer
is the building or tree mask, for heights and the building volume the object height on building or tree pixels, and for slope order the slope of the terrain model. For line of sight it is the terrain,
the surface and the sight line between the two eyes along the line.

\section{GeoDSL Reference and Prompt}
\label{sec:supp_dsl}

\noindent\textbf{Layers and functions.}
Table~\ref{tab:supp_dsl} lists the vocabulary of GeoDSL. A program is exactly one answer constructor applied to a
score, which is an arithmetic expression over layers, functions and numbers in the option variable \texttt{o}. The
regions, points or point pair of a question are parsed from its text by a fixed rule, so the program only states what
to compute. The static checker infers whether each subexpression is a layer, a number, a condition or an option,
checks that a function receives a region or a point where it requires one, and allows comparisons only inside
\texttt{ifelse} and \texttt{yes\_no}.

\begin{table}[h]
\centering
\small
\setlength{\tabcolsep}{6pt}
\renewcommand{\arraystretch}{1.08}
\begin{tabular}{@{}>{\raggedright\arraybackslash}p{5.0cm}>{\raggedright\arraybackslash}p{11.0cm}@{}}
\toprule
Name & Meaning \\
\midrule
\multicolumn{2}{@{}l}{\textit{Layers}} \\
\texttt{OBJECT\_HEIGHT} & height of buildings and trees above the ground, m, from the height sensor \\
\texttt{HEIGHT} & surface height above sea level, m, ground plus objects \\
\texttt{GROUND} & bare ground above sea level, m, from Eq.~\eqref{eq:ground} \\
\texttt{SLOPE} & slope of \texttt{GROUND} after a $5\times5$ mean filter, degrees \\
\texttt{SVF} & sky view factor from 0 to 1, computed with UMEP from \texttt{OBJECT\_HEIGHT} \\
\texttt{BUILDING} & building mask from the segmentation head \\
\texttt{TREE} & tree mask from the segmentation head \\
\texttt{NATURE} & mask of natural cover from the segmentation head \\
\texttt{EDGES} & edge map of the image \\
\texttt{GRADIENT} & gradient magnitude of the image \\
\texttt{BRIGHTNESS} & grey level of the image \\
\midrule
\multicolumn{2}{@{}l}{\textit{Measurement functions}} \\
\texttt{mean(L, o)} & mean of L over region o, or its value at point o \\
\texttt{std(L, o)}, \texttt{var(L, o)} & standard deviation and variance of L over region o \\
\texttt{at(L, o)} & value of L at point o \\
\texttt{window\_mean(L, o, r)}, \texttt{window\_std(L, o, r)} & mean and standard deviation of L in the
$(2r+1)^2$ window around point o \\
\texttt{masked\_mean(L, M, o)} & mean of L over the pixels of region o where mask M is set, 0 if there are none \\
\texttt{count(M, o)} & separate objects of M of at least 20\,m$^2$ lying at least half inside region o \\
\texttt{object\_max(L, M, o)} & largest per-object median of L over objects of M of at least 50\,m$^2$ lying at
least half inside region o, 0 if there are none \\
\texttt{viewshed(o)} & mean free line-of-sight length in eight directions from point o \\
\texttt{clearance(x)} & how far anything rises above the sight line between the two points of the question, with
eyes $x$\,m above the ground \\
\texttt{exp}, \texttt{abs}, \texttt{min}, \texttt{max} & arithmetic helpers \\
\texttt{ifelse(c, a, b)} & $a$ if the condition holds, else $b$ \\
\midrule
\multicolumn{2}{@{}l}{\textit{Answer constructors}} \\
\texttt{pick\_max(s)}, \texttt{pick\_min(s)} & the option with the largest or smallest score \\
\texttt{rank\_desc(s)}, \texttt{rank\_asc(s)} & all options ordered by score \\
\texttt{value(s, step, unit)} & the score of the target region or point, rounded to a multiple of \texttt{step} \\
\texttt{yes\_no(c)} & Yes if the condition holds \\
\texttt{top\_classes(k)}, \texttt{classes\_present} & land-cover classes of the whole tile \\
\bottomrule
\end{tabular}
\caption{\textbf{GeoDSL vocabulary.} L is a layer, M a mask layer, o the option variable, s a score and c a
condition.}
\label{tab:supp_dsl}
\end{table}

\noindent\textbf{Prompt.}
The writer receives the system prompt below, a user message, shown for the question of Fig.~\ref{fig:teaser}, and
execution feedback when a program fails. The grammar lists the constructors, layers and functions of Table~\ref{tab:supp_dsl} with their
signatures and gives one example, which picks the region with the tallest buildings on average. No benchmark asks this
choice, and \bench{} asks for the same quantity as a number. The metric notes transcribe the scoring formulas of the
Geo3DVQA paper, with the conventions it leaves open, such as window sizes, chosen on its development set, and are
used unchanged on \bench{}. When the checker or a dry run on a synthetic tile rejects a
program, the feedback message returns the error, and the writer has two rounds to repair its program. Qwen3-4B runs in vLLM with a context of 8,192 tokens, at most 512 output tokens, temperature 0 and
seed 0.

\begin{promptbox}{System prompt (complete)}
\begin{verbatim}
You write programs in GeoDSL that answer questions about one aerial image tile of 1000 x 1000 pixels
    (1 m per pixel).
The regions or candidate points named in the question are already parsed for you. The variable `o`
    stands for one option
(or for the single target region). Your program says how to score `o`; an executor evaluates it on
    the tile's data layers
and turns the scores into the answer.

A program is exactly one call:
  pick_max(score)  -- the option with the largest score
  pick_min(score)  -- the option with the smallest score
  rank_desc(score)  -- all options ordered from largest to smallest score
  rank_asc(score)  -- all options ordered from smallest to largest score
  value(score, step=..., unit=...)  -- the score of the target region or target point rounded to a
      multiple of `step`: step=0.1 gives one decimal place (0.7), step=10 gives tens (70); keyword
      `unit` is appended, e.g. unit="m"
  yes_no(score)  -- Yes if the condition holds, else No, e.g. yes_no(x < 0)
  top_classes(score)  -- the k land-cover classes with the largest area in the whole tile, e.g.
      top_classes(k=3)
  classes_present(score)  -- every land-cover class covering more than min_share of the whole tile,
      e.g. classes_present(min_share=0)

`score` is an arithmetic expression (+ - * / **, numbers, parentheses) built from layers and
    functions.
Layers:
  SVF        sky view factor, 0..1
  OBJECT_HEIGHT height of objects (buildings, trees) above the ground (nDSM), metres
  HEIGHT     absolute height of the surface above sea level (DSM), metres: what the questions call
      the height or elevation of an area; also used for lines of sight
  BUILDING   building mask, 1 on building pixels else 0
  NATURE     natural cover mask (forest, water, agriculture, grass), 1 / 0
  EDGES      binary edge map of the RGB image (Canny), 1 / 0
  GRADIENT   RGB gradient magnitude (Sobel), 0..~1000
  BRIGHTNESS RGB grey level, 0..1
  TREE       tree mask, 1 on tree pixels else 0
  GROUND     height of the bare ground above sea level (HEIGHT minus OBJECT_HEIGHT), metres
  SLOPE      slope of the bare ground (GROUND smoothed over 5 x 5 m), degrees
Functions:
  mean(LAYER, o)  -- mean of a layer over region o (value at the pixel if o is a point)
  std(LAYER, o)  -- standard deviation of a layer over region o [o must be a region]
  var(LAYER, o)  -- variance of a layer over region o [o must be a region]
  at(LAYER, o)  -- value of a layer at point o [o must be a point]
  window_mean(LAYER, o, r)  -- mean of a layer in the (2r+1)^2 window around point o [o must be a
      point]
  window_std(LAYER, o, r)  -- std of a layer in the (2r+1)^2 window around point o [o must be a
      point]
  masked_mean(LAYER, MASK, o)  -- mean of the first layer over pixels of region o where the mask
      layer is 1 (0 if none) [o must be a region]
  viewshed(o, eye=2.0, max_dist=100.0)  -- mean free line-of-sight length over 8 directions from
      point o, as a fraction of max_dist px [o must be a point]
  exp(x)  -- e**x
  abs(x)  -- |x|
  min(a, b)  -- smaller of two numbers
  max(a, b)  -- larger of two numbers
  ifelse(cond, a, b)  -- a if cond else b
  count(LAYER, o)  -- number of separate objects of a mask layer (connected blobs of at least 20
      m^2) lying at least half inside region o [o must be a region]
  object_max(LAYER, LAYER, o)  -- over the objects of the mask (second layer; blobs of at least 50
      m^2, at least half inside region o), the largest per-object median of the first layer; 0 if
      there is none [o must be a region]
  clearance(x)  -- how far (m) anything rises above the straight sight line between the two points
      of the question, with eyes x m above the ground at both ends; above 0 means the view is
      blocked [o must be a pair]
Comparisons (<, >, <=, >=) are only allowed as the first argument of ifelse or yes_no.
Use only these names. No Python statements, variables, lambdas or comments.

Example
Question: Which region has the tallest buildings on average?
A: [xmin=10%, ymin=10%, xmax=30%, ymax=30%]
B: [xmin=60%, ymin=20%, xmax=80%, ymax=40%]
Context: OPTIONS = 2 rectangular regions (A, B); o is a region
<program>pick_max(masked_mean(OBJECT_HEIGHT, BUILDING, o))</program>

Metric dictionary (the definitions this benchmark uses; apply them when a question names a metric):
- Sky View Factor (SVF) of a region: mean(SVF, o); of a point: at(SVF, o)
- SVF variability of a region (standard deviation): std(SVF, o); "most consistent" means the lowest
    std
- spatial openness index: (mean(SVF, o) + 0.5 * std(SVF, o)) / 1.25
- low building density: 1 - mean(BUILDING, o)
- terrain flatness: exp(-std(HEIGHT, o) / 5)
- visual simplicity: exp(-var(GRADIENT, o) / 500)
- penalty for nearby buildings (around a point): window_mean(BUILDING, o, 4)
- Building Coverage Ratio (share of the region covered by buildings): mean(BUILDING, o)  (not
    masked_mean)
- Floor Area Ratio: mean(BUILDING, o) * min(20, masked_mean(OBJECT_HEIGHT, BUILDING, o) / 3.5) / 5
- Sky View Factor penalty (urban density): ifelse(mean(NATURE, o) > 0.5, 0, 1 - mean(SVF, o))
- Edge Density: mean(EDGES, o)
- Brightness analysis (urban density; darker means denser): 1 - mean(BRIGHTNESS, o)
- viewshed distance: viewshed(o, eye=2.0, max_dist=100)
- terrain roughness variation: exp(-(window_std(HEIGHT, o, 5) - 10) ** 2 / 200)
- the "height" or "elevation" of an area or region: mean(HEIGHT, o) (absolute, above sea level);
  OBJECT_HEIGHT is only for the height of buildings or trees above the ground
- A score "X (50%), Y (25%), ..." is 0.5 * X + 0.25 * Y + ...; a penalty with weight w is subtracted
    (- w * penalty).

If the question asks for one option, use pick_max or pick_min; use rank_desc / rank_asc only when it
    asks to order all options.
Reply with the program only, inside <program></program>.
\end{verbatim}
\end{promptbox}

\begin{promptbox}{User message (abridged)}
\begin{verbatim}
Question: Identify the area that encompasses the greatest volume of constructed edifices
from the list provided.

A: [xmin=89%, ymin=72%, xmax=99%, ymax=82%]
B: [xmin=75%, ymin=49%, xmax=85%, ymax=59%]
C: [xmin=62%, ymin=70%, xmax=72%, ymax=80%]
D: [xmin=8%, ymin=13%, xmax=18%, ymax=23%]
...
Context: OPTIONS = 4 rectangular regions (A, B, C, D); o is a region; the answer is ONE
option
\end{verbatim}
\end{promptbox}

\begin{promptbox}{Execution feedback, sent only when a program fails}
\begin{verbatim}
The program failed: yes_no() needs a comparison such as x < 0
Return a corrected program inside <program></program>.
\end{verbatim}
\end{promptbox}

\noindent\begin{minipage}{\linewidth}
\begin{promptbox}{Programs written by the frozen writer for \bench{}-CH questions}
\begin{verbatim}
Mean building height      value(masked_mean(OBJECT_HEIGHT, BUILDING, o), step=1, unit="m")
Tallest building          value(object_max(OBJECT_HEIGHT, BUILDING, o), step=1, unit="m")
Largest building volume   pick_max(mean(BUILDING, o) * masked_mean(OBJECT_HEIGHT, BUILDING, o))
Lowest ground             pick_min(mean(GROUND, o))
Line of sight             yes_no(clearance(1.6) <= 0)
\end{verbatim}
\end{promptbox}
\end{minipage}

\section{Baseline Details}
\label{sec:supp_baselines}

\noindent\textbf{Image-only VLMs.}
Each VLM receives the image and the question followed by the instruction below, the protocol of Geo3DVQA. Decoding is
greedy where a model allows it, with a limit of 4,096 output tokens through APIs and 1,024 for local models, 512 for
Qwen3-VL-8B on Geo3DVQA, and reasoning is switched off where a model allows it. GPT-6.1 Terra runs with reasoning effort none and temperature 0. Gemini 3.8 Flash cannot switch reasoning off and runs at its lowest effort. Local
general-purpose models run with vLLM~\cite{vllm}, GeoChat and EarthDial with their official inference code, and
SpaceThinker with Transformers.

\begin{promptbox}{Instruction appended to every question for image-only VLMs}
\begin{verbatim}
Please analyze the provided image(s) and answer the question. After your analysis,
provide your final answer in the format:
<final short answer>your answer</final short answer>
\end{verbatim}
\end{promptbox}

\noindent\textbf{Normalisation.}
All baselines share one lenient answer extraction before the scorer. It reads the requested tag pair or, failing
that, the usual answer markers and the last line, strips markup and special tokens, and maps a lone option letter to
its region, but never changes a number.

\noindent\textbf{Fine-tuned models.}
Both fine-tuned baselines follow the LoRA recipe of Geo3DVQA on Qwen2.5-VL-7B-Instruct, with the vision tower frozen.
The Geo3DVQA model is trained on the released 100K training questions plus 980 free-form questions, and the in-domain
model with the same settings on the 3,357 \bench{}-CH training questions. We report both with their training format. With the image-only prompt, the Geo3DVQA model reaches 30.0\%
on \bench{}.

\noindent\textbf{Measurements given to VLMs.}
For each option of a question, the measurement text lists the quantities that GeoDSL functions compute on the same
predicted layers and ground. Points receive the corresponding point values and the clearance of their sight line.
The text is placed between the image and the question. Qwen3-4B, our writer, receives the same text and instruction without the image, with ``image(s)'' replaced by ``measurements'', and decodes greedily with at most 4,096 output tokens. For the question of Fig.~\ref{fig:teaser} it begins as below.

\begin{promptbox}{Measurement text given to VLMs (abridged)}
\begin{verbatim}
Measured on predicted height, building, tree and sky view factor maps (ground elevation
is from a global 30 m elevation model): Region A: mean SVF 0.798, mean object height
2.3 m, building share 16.5%, tree share 15.2%, number of buildings 7, tallest building
7.5 m, mean building height 5.7 m, mean tree height 5.7 m, mean ground elevation 505.3 m
above sea level, mean ground slope 6.2 degrees; Region B: ...
\end{verbatim}
\end{promptbox}

In the heights-only condition of the main paper, the same text keeps only the quantities in metres of height,
namely mean object height, tallest building, mean building and tree height and ground elevation. It drops sky view,
shares, counts, slope, the class under a point and the sight-line clearance.

In the height-map condition, the image is followed by two images of its size, the predicted object heights above
ground and the ground elevation above sea level that \method{} reads. Both use the viridis colour map with a colour
bar in metres scaled to the tile. A short text names the three images before the question.

\noindent\textbf{Visual programs.}
The VisProg-style and ViperGPT-style baselines use the frozen 4B writer with greedy decoding, the same question
text and metric notes, the same layers and functions and the same sensors, and change only the program form. The
ViperGPT-style writer produces a Python function in which the GeoDSL functions are callables and math and NumPy can be
imported, and it forms the answer string itself. Failing code is returned with its error for up to two repairs. The VisProg-style writer produces one
module call per line, with expression lines for arithmetic and comparisons and a final result line, following
in-context examples that we wrote on the Swiss development set. Neither form is type-checked before execution.
Their complete prompts, and the model identifiers and decoding settings of every API run, are part of the
released code.

\section{Street-Level Transfer}
\label{sec:supp_street}

\noindent\textbf{Split and answers.}
SpatialRGPT-Bench has 1,406 questions on 688 frames, with the regions given as masks. A random split by frame keeps
172 frames with 388 questions for development and 516 frames with 1,018 questions for testing. Distances and sizes are converted to metres, directions to clock hours, and
qualitative answers to yes or no or to the region they name.

\noindent\textbf{Sensor.}
Depth Pro~\cite{depthpro} predicts metric depth and its own focal length, from which every pixel is back-projected
into a frame with $x$ to the right, $y$ up and $z$ away from the viewer. A region is the set of points inside its
mask.

\noindent\textbf{SceneDSL.}
Programs refer to the two regions or, inside \texttt{pick\_max} and \texttt{pick\_min}, to the option \texttt{o}.
Seven functions measure one region, namely \texttt{width}, \texttt{height}, \texttt{depth}, \texttt{size},
\texttt{left\_right}, \texttt{elevation} and \texttt{distance\_from\_viewer}, and three measure a pair, namely
\texttt{distance}, \texttt{horizontal\_distance} and \texttt{vertical\_distance}. The constructors are
\texttt{yes\_no}, \texttt{pick\_max}, \texttt{pick\_min}, \texttt{value} and \texttt{clock}. A region's centre is the
median of its points and its box spans the 2nd to the 98th percentile along each axis. Width, height and depth are the
box extents along $x$, $y$ and $z$, and size is their product. Distances are taken between centres, and the horizontal and
vertical distances are the differences of the centres along $x$ and $y$. The distance from the viewer is the depth of
the centre, and the clock direction from A to B is $12-\lfloor\theta/30^\circ\rfloor$ with
$\theta=\operatorname{atan2}(-\Delta x,\Delta z)$, as in the benchmark's generator. We fixed these conventions on the
development split. Size has degree 3 in the depth scale and every other function degree 1, and the scale-invariance check follows the rules of Sec.~\ref{sec:cert}. It marks 565 test answers, all of which read the
depth. The writer and its repair rounds are those of \method{}, with a
prompt that lists the SceneDSL vocabulary, a context of 4,096 tokens and at most 256 output tokens.

\noindent\textbf{Scoring.}
As in SpatialRGPT-Bench, a distance or size is correct within $[0.75, 1.25]$ times the reference, a clock direction
within one hour, and a qualitative answer when it matches. We score deterministically without a language-model judge.
The published numbers marked $^\dagger$ in Table~\ref{tab:street} use all 1,406 questions and a GPT-4 judge and are
therefore not strictly comparable. Per type, \method{} is 5 to 11 points above the published SpatialReasoner-R1 on the
three distance types and 15 to 34 points below it on widths and heights.

\section{Writer Ablation}
\label{sec:supp_writers}

\begin{table}[h]
\centering
\small
\setlength{\tabcolsep}{6pt}
\begin{tabular}{@{}lrrrrr@{}}
\toprule
& \multicolumn{2}{c}{\bench{}-CH} & \multicolumn{2}{c}{\bench{}-FR} & \\
\cmidrule(lr){2-3}\cmidrule(lr){4-5}
Writer & Accuracy & Invalid & Accuracy & Invalid & \method{} $-$ writer, pooled \\
\midrule
Qwen3-4B (main) & 72.4 & 3 & 72.7 & 6 & -- \\
\midrule
Qwen3-8B & 56.0 & 131 & 57.1 & 132 & \pp{+16.0} \\
Qwen3-14B & 60.5 & 6 & 63.1 & 7 & \pp{+10.8} \\
Qwen3-30B-A3B-2507 & 57.5 & 165 & 60.5 & 158 & \pp{+13.6} \\
Qwen3.5-4B & 58.5 & 189 & 63.0 & 149 & \pp{+11.8} \\
Qwen3.5-9B & 67.1 & 69 & 65.2 & 92 & \pp{+6.4} \\
Qwen3.5-27B & \textbf{73.9} & 1 & \underline{75.7} & 2 & \pp{-2.2} \\
DeepSeek-V4.1-Flash & \underline{73.8} & 0 & \textbf{76.0} & 0 & \pp{-2.3} \\
DeepSeek-V4-Pro & 67.9 & 0 & 69.8 & 1 & \pp{+3.7} \\
\bottomrule
\end{tabular}
\caption{\textbf{Writers.} Every writer receives the same prompt with execution feedback; the language, checker,
executor and sensors are frozen. Invalid counts programs that still fail after two repairs. Bold and underline
mark the best two per column.}
\label{tab:supp_writers}
\end{table}

Table~\ref{tab:supp_writers} lists all eight writers we swapped in. The prompt was written with Qwen3-4B on the
development splits and is not adapted to the other writers. Thinking models write with thinking switched off,
as the frozen writer does. Two writers with several times the parameters of ours, Qwen3.5-27B and DeepSeek-V4.1-Flash,
are 1.4 to 3.3 points better in both countries, and the paired differences are significant. Qwen3-8B,
Qwen3-30B-A3B, Qwen3.5-4B and Qwen3.5-9B lose through 161 to 338 invalid programs, while Qwen3-14B and DeepSeek-V4-Pro
write valid programs that compute other quantities.

\begin{table}[h]
\centering
\small
\setlength{\tabcolsep}{4pt}
\begin{tabular}{@{}llrrrrr@{}}
\toprule
& & \multicolumn{3}{c}{Accuracy} & & \\
\cmidrule(lr){3-5}
Writer & Program form & CH & FR & Pooled & Runs & Ref. \\
\midrule
Qwen3-4B & GeoDSL (\method{}) & \textbf{72.4} & \textbf{72.7} & \textbf{72.6} & \textbf{99.7} & \textbf{94.3} \\
& VisProg-style & 58.7 & 60.0 & 59.4 & 85.6 & 78.2 \\
& ViperGPT-style & 55.8 & 56.6 & 56.2 & 85.5 & 76.5 \\
\midrule
DeepSeek-V4.1-Flash & GeoDSL & 73.8 & 76.0 & 74.9 & \textbf{100.0} & 98.9 \\
& VisProg-style & 67.6 & 69.1 & 68.3 & \textbf{100.0} & 92.1 \\
& ViperGPT-style & \textbf{74.4} & \textbf{76.1} & \textbf{75.3} & 99.7 & \textbf{99.7} \\
\bottomrule
\end{tabular}
\caption{\textbf{Program forms with two writers} on the \bench{} test sets (\%). Sensors, functions, metric notes
and repair rounds are the same in every row. Runs is the share of programs that return an answer, and Ref.\ the
share whose answer equals that of the type's most frequent development program on the same predicted maps. Bold
marks the best form per writer.}
\label{tab:supp_forms}
\end{table}

Table~\ref{tab:supp_forms} repeats the comparison of program forms with DeepSeek-V4.1-Flash, one of the two writers
that score above ours in GeoDSL. With this writer, free Python and GeoDSL are level, with a pooled difference of
$-0.4$ [$-0.8$, $+0.1$]. DeepSeek-V4.1-Flash writes a GeoDSL program in 22 output tokens per question, free Python in 59 and a step
list in 100. Moving to our 4B writer costs free Python 19.1 points, the step lists 9.0 and GeoDSL 2.3. Free-form answers that differ from the reference are
correct in 1 to 14\% of cases, so the Ref.\ column counts wrong quantities rather than valid alternatives.

\section{Per-Type Results}
\label{sec:supp_types}

\begin{table}[h]
\centering
\footnotesize
\setlength{\tabcolsep}{2.6pt}
\begin{tabular}{@{}lrr*{11}{r}@{}}
\toprule
& & & & \multicolumn{4}{c}{Image only} & & \multicolumn{2}{c}{Programs} & & & \\
\cmidrule(lr){5-8}\cmidrule(lr){10-11}
Type & $n$ & Inv. & Const. & \shortstack[r]{Gemma\,4\\31B} & \shortstack[r]{Qwen3.7\\Plus} & \shortstack[r]{Gemini\\3.8 Fl.}
& \shortstack[r]{GPT-6.1\\Terra} & \shortstack[r]{FT in\\domain} & VisProg & \shortstack[r]{Viper\\GPT}
& \shortstack[r]{GPT +\\meas.} & Ours & \shortstack[r]{Ours,\\ref.} \\
\midrule
Mean building height & 161 & 0 & 58.4 & 62.7 & 62.1 & 79.5 & 73.9 & 55.3 & 79.5 & 82.0 & \textbf{87.6} & \textbf{87.6} & 100.0 \\
Tallest building & 210 & 0 & 48.1 & 51.4 & 46.7 & \textbf{65.2} & 57.1 & 41.9 & 55.7 & 61.4 & 64.8 & 64.8 & 87.6 \\
Mean tree height & 238 & 41 & 45.8 & 35.3 & 42.4 & \textbf{67.6} & 55.9 & 38.7 & 27.7 & 32.8 & 36.1 & 19.3 & 55.0 \\
Tree cover & 240 & 100 & 23.8 & 28.8 & 33.3 & 58.8 & 66.7 & 40.0 & 71.2 & 15.8 & \textbf{76.7} & 74.6 & 87.9 \\
Building cover & 120 & 100 & 49.2 & 61.7 & 73.3 & 90.0 & 84.2 & 70.0 & \textbf{95.8} & 87.5 & \textbf{95.8} & \textbf{95.8} & 100.0 \\
Most built-up region & 89 & 100 & 25.8 & 76.4 & 85.4 & 89.9 & 93.3 & 60.7 & 77.5 & \textbf{98.9} & \textbf{98.9} & \textbf{98.9} & 100.0 \\
Number of buildings & 205 & 96 & 37.1 & 22.0 & 24.9 & 35.6 & 36.6 & 41.0 & \textbf{72.7} & \textbf{72.7} & 64.9 & 69.8 & 95.6 \\
Largest building volume & 218 & 52 & 25.2 & 63.3 & 66.5 & 81.7 & 81.2 & 53.7 & 50.0 & 15.1 & 91.7 & \textbf{93.6} & 100.0 \\
Sky view factor at a point & 240 & 0 & 19.2 & 24.6 & 24.6 & 27.9 & 30.4 & 21.7 & 25.0 & \textbf{43.3} & \textbf{43.3} & \textbf{43.3} & 100.0 \\
Most open sky & 239 & 0 & 25.1 & 61.1 & 71.1 & 81.6 & 83.3 & 54.4 & 86.6 & 86.6 & \textbf{87.0} & 86.6 & 100.0 \\
Best roof for solar panels & 180 & 0 & 25.0 & 30.6 & 33.9 & 40.6 & 41.7 & 29.4 & 49.4 & 49.4 & \textbf{50.0} & 49.4 & 62.2 \\
Lowest ground & 229 & 0 & 25.3 & 43.2 & 46.7 & 60.3 & 56.8 & 38.0 & 95.6 & 66.8 & \textbf{98.7} & 98.3 & 99.6 \\
Order by slope & 223 & 2 & 17.0 & 45.3 & 39.9 & 57.8 & 55.2 & 20.6 & \textbf{74.0} & 48.0 & \textbf{74.0} & 67.7 & 89.7 \\
Ground elevation & 60 & 0 & 23.3 & 28.3 & 16.7 & 45.0 & 25.0 & 10.0 & 70.0 & 95.0 & 95.0 & \textbf{96.7} & 98.3 \\
Line of sight & 237 & 0 & 49.4 & 58.2 & 61.6 & 66.7 & 65.8 & 48.9 & 3.8 & 65.0 & \textbf{89.0} & 88.6 & 98.7 \\
\midrule
All & 2,889 & 30 & 33.0 & 45.1 & 47.8 & 62.1 & 60.2 & 41.3 & 59.4 & 56.2 & \textbf{74.2} & 72.6 & 90.8 \\
\bottomrule
\end{tabular}
\caption{\textbf{Accuracy (\%) per question type,} pooled over \bench{}-CH and -FR. Inv.\ is the share of
\method{}'s answers marked as invariant to the height scale. Bold marks the best system per type, excluding
\method{} on reference maps.}
\label{tab:supp_types}
\end{table}

Table~\ref{tab:supp_types} breaks the pooled accuracy down by type. The systems that read our sensors lead on
every type except the tallest building and the mean tree height, and among them \method{} is within one point of the
best on 11 of the 15 types. The largest gaps to the image-only models lie in ground elevation, lowest ground,
counts, line of sight and volumes. Mean tree height is \method{}'s weakest type at 19.3\%, and both the writer and the
sensor contribute. Of its programs, 55\% take the mean of \dsl{OBJECT\_HEIGHT} over \dsl{TREE}, and the others average
the tree mask or scale the tree cover, which limits the reference maps to 55.0\%. With the right program, 35\% of the
answers are correct on predicted maps and all of them on reference maps. The height sensor returns 73\% of the
reference tree height at the median and half for trees above 20\,m. The sky view factor at a point remains hard for all systems, at 43\% at
best. Every type gains on the reference maps. On line of sight, the VisProg-style step lists rarely compute a
clearance.

\noindent\textbf{Volumes.}
The volume program multiplies building cover by mean building height, which is the volume per unit area. The four
regions of every volume question have the same size, so this orders them exactly like total volume.

\noindent\textbf{Geo3DVQA.}
Table~\ref{tab:supp_g3d} breaks the Geo3DVQA accuracy down by category. On each of the ten categories that a program
computes, \method{} is the best system or level with another program form over the same sensors. Its largest leads
over the fine-tuned model lie in height average and highest region, which ask for elevation above sea level and need
the ground that the elevation model supplies. GPT-6.1 Terra with the image alone leads the image-only models,
including the fine-tuned one, on eight of the 12 categories. The fine-tuned model leads on land use and land-cover
type, where \method{} returns the most frequent training answer. The VisProg-style step lists return no answer on all
but one of the openness, building density and visibility range questions.

\begin{table}[h]
\centering
\footnotesize
\setlength{\tabcolsep}{3pt}
\begin{tabular}{@{}lrrrrrrrr@{}}
\toprule
& & & & \multicolumn{2}{c}{Image only} & \multicolumn{3}{c}{Programs, our sensors} \\
\cmidrule(lr){5-6}\cmidrule(lr){7-9}
Category & $n$ & Const. & \shortstack[r]{FT 7B,\\repr.} & \shortstack[r]{Qwen3.6\\27B} & \shortstack[r]{GPT-6.1\\Terra}
& VisProg & \shortstack[r]{Viper\\GPT} & Ours \\
\midrule
SVF value & 999 & 24.8 & 46.5 & 22.4 & 46.8 & 55.1 & 55.3 & \textbf{60.3} \\
Region ranking & 999 & 17.2 & 51.6 & 55.9 & 82.0 & \textbf{86.5} & 74.9 & \textbf{86.5} \\
Regional SVF variability & 662 & 34.7 & 57.4 & 40.8 & 55.3 & \textbf{78.7} & \textbf{78.7} & \textbf{78.7} \\
Sun exposure & 999 & 24.5 & 51.4 & 48.8 & 66.0 & \textbf{76.5} & \textbf{76.5} & \textbf{76.5} \\
Height average & 666 & 35.4 & 44.0 & 11.0 & 15.6 & 29.6 & 91.7 & \textbf{98.9} \\
Highest region & 666 & 26.3 & 42.8 & 37.1 & 47.7 & \textbf{86.9} & \textbf{86.9} & \textbf{86.9} \\
Land use & 666 & 16.2 & \textbf{59.6} & 18.0 & 14.0 & -- & -- & 16.2 \\
Land-cover type & 993 & 71.9 & \textbf{88.9} & 51.9 & 61.3 & -- & -- & 71.9 \\
Openness & 999 & 24.1 & 62.8 & 55.7 & 79.0 & 0.0 & 89.5 & \textbf{89.9} \\
Sky visibility & 999 & 24.1 & 63.4 & 50.4 & 71.2 & 21.6 & 76.9 & \textbf{77.4} \\
Building density & 585 & 26.3 & 53.8 & 51.1 & 69.4 & 0.0 & 82.7 & \textbf{83.1} \\
Visibility range & 999 & 23.2 & 38.4 & 37.2 & 44.7 & 0.0 & 65.8 & \textbf{66.1} \\
\midrule
Ten program categories & 8,573 & 25.4 & 51.4 & 41.9 & 59.3 & 43.1 & 76.7 & \textbf{79.4} \\
All 12 categories & 10,232 & 29.3 & 55.6 & 41.3 & 56.6 & -- & -- & \textbf{74.6} \\
\bottomrule
\end{tabular}
\caption{\textbf{Accuracy (\%) per category on the Geo3DVQA test set.} FT 7B is our reproduction of the
fine-tuned Qwen2.5-VL-7B. The program forms cover the ten program categories. Bold marks the best system per
category, excluding the constant.}
\label{tab:supp_g3d}
\end{table}

\noindent\textbf{New compositions.}
To test whether the writer composes functions beyond the 15 types, we drew 379 questions of four new types on the
120 test tiles of both countries, at most one per type and tile, with the reference layers, region sampler and
margins of \bench{}. They ask for the region with the
largest tree volume, the largest building cover minus tree cover and the highest ratio of building to tree height, and
for an ordering by sky view factor minus 0.02 times the slope. No prompt, metric note or development question covers
them. Table~\ref{tab:supp_compose} gives the accuracy and the share of programs that compute the defined quantity,
against 25\% by chance for the choices and 17\% for the orderings. We judged every distinct program against the
definition, counting algebraically equal forms and small guard terms such as $+10^{-6}$ as correct.
\method{} answers 83.1\% \ci{79.5}{86.7} on predicted maps.

\begin{table}[h]
\centering
\small
\setlength{\tabcolsep}{6pt}
\begin{tabular}{@{}lrccc@{}}
\toprule
New type & $n$ & Predicted maps & Reference maps & Defined program \\
\midrule
Largest tree volume & 120 & 74.2 & 85.8 & 43 / 120 \\
Building cover minus tree cover & 120 & 95.0 & 100.0 & 120 / 120 \\
Building to tree height ratio & 20 & 80.0 & 70.0 & 6 / 20 \\
Order by sky view minus slope & 119 & 80.7 & 100.0 & 119 / 119 \\
\midrule
All & 379 & 83.1 & 93.9 & 288 / 379 \\
\bottomrule
\end{tabular}
\caption{\textbf{New composite types} on all test tiles of both countries: accuracy (\%) and programs that compute
the defined quantity.}
\label{tab:supp_compose}
\end{table}

The writer composes the defined program for every cover difference and every ordering. The other tree-volume programs take the mean tree height without the
cover, and the other ratio programs divide by the mean of the tree mask, as for mean tree height. Both still pick
the right region on most reference maps, since tree height correlates with tree volume and building height with the
height ratio. Each of the 20 ratio questions is worth 5 points.

\section{Sensor Testbed and Scale Check}
\label{sec:supp_sensors}

\begin{table}[h]
\centering
\small
\setlength{\tabcolsep}{2.4pt}
\begin{tabular}{@{}lrrrrrrrr@{}}
\toprule
& \multicolumn{4}{c}{\bench{}-CH} & \multicolumn{4}{c}{\bench{}-FR} \\
\cmidrule(lr){2-5}\cmidrule(l){6-9}
& \multicolumn{2}{c}{Invariant} & & & \multicolumn{2}{c}{Invariant} & & \\
\cmidrule(lr){2-3}\cmidrule(lr){6-7}
Height estimator & No $h$ & $h$ & Other & All & No $h$ & $h$ & Other & All \\
\midrule
AltiMap v2 (main) & 73.0 & 91.8 & 71.1 & 72.4 & 68.0 & 94.2 & 73.3 & 72.7 \\
AltiMap v2, coarser input & 73.0 & 91.8 & 68.3 & 70.5 & 68.0 & 96.2 & 70.0 & 70.4 \\
AltiMap v1 & 73.0 & 95.1 & 58.4 & 63.7 & 68.0 & 92.3 & 58.5 & 62.2 \\
RS3DAda & 73.0 & 88.5 & 50.8 & 58.1 & 68.0 & 92.3 & 52.1 & 57.7 \\
HighResCanopyHeight & 73.0 & 73.8 & 50.0 & 56.9 & 68.0 & 71.2 & 52.8 & 57.4 \\
\bottomrule
\end{tabular}
\caption{\textbf{Height estimators under the same programs (\%).} Invariant answers are split into those that read no
height (No $h$; 385 in Switzerland and 362 in France) and those that read heights ($h$; 61 and 52, all volume
comparisons). Other answers are the 1,037 and 992 that the check does not mark.}
\label{tab:supp_sens}
\end{table}

\noindent\textbf{Sensor settings.}
AltiMap v2 reads the image upsampled by 4 for heights and by 3, to its training resolution of 0.33\,m, for the
segmentation, whose building and tree classes give the two masks. Both outputs are averaged back to 1\,m.

\noindent\textbf{Testbed.}
We run the programs of \method{} with five height estimators, namely AltiMap v2 at our input scale ($4\times$) and at a
coarser one ($3\times$), AltiMap v1, RS3DAda~\cite{synrs3d} and HighResCanopyHeight~\cite{hrch}. The segmentation, the elevation model and the
programs stay fixed, while the ground of Eq.~\eqref{eq:ground} and the sky view factors are recomputed from each
estimate, as in the full pipeline. Table~\ref{tab:supp_sens} splits each run by the scale-invariance check and, within the invariant answers, by whether
the program reads the height layer. The invariant answers that read no height are tree and building covers, building
counts and the most built-up region. They also include 102 wrong programs that read no height either, 98 of them
tree-height programs that average or scale the tree mask and answer in metres. Their accuracy is the same under every
estimator. The invariant answers that read heights are the volume comparisons. Across the first four estimators their accuracy varies by 7 and 4 points in Switzerland and France,
against 20 and 21 points for the other answers, which report metres or read sky view, ground, slope or sight lines
that mix heights with other quantities. The canopy height model lowers both groups by 18 to 23 points.

\noindent\textbf{Sensor errors.}
On the object pixels of the test tiles, AltiMap v2 underestimates heights, with a median per-tile factor of 0.60 in
Switzerland and 0.75 in France. One factor per tile removes 69\% and 51\% of the squared height error. A large part of
the error is therefore a stretch of
the whole tile, which the check covers, and the rest varies within a tile, for instance between trees and
buildings. Against the LiDAR terrain models, GLO-30 alone lies 4.7 and 3.0\,m above the ground on average, and removing the
predicted objects with Eq.~\eqref{eq:ground} brings this bias to $+0.5$ and $-1.5$\,m. On the development sets, $\beta = 0$, 0.5 and 1 give accuracies within 0.2 points on
\bench{}-CH and within 1.1 points on Geo3DVQA.

\noindent\textbf{Agreement proxy.}
For each estimator we also count how often the answers on its maps agree with the answers of the same programs on
the reference maps. This needs reference heights but no answer labels. The agreement ranks the five estimators
exactly as test accuracy does on \bench{}-CH, with agreement from 52.5\% to 62.5\%, and on \bench{}-FR up to one swap
of the last two estimators, whose accuracies differ by 0.3 points. Over 10,000 random draws of twenty test tiles, the
estimator with the highest agreement on them is the most accurate one in 97\% and 88\% of draws in Switzerland and
France.

\section{Run Time}
\label{sec:supp_cost}

\begin{figure}[!ht]
\centering
\includegraphics[width=\linewidth]{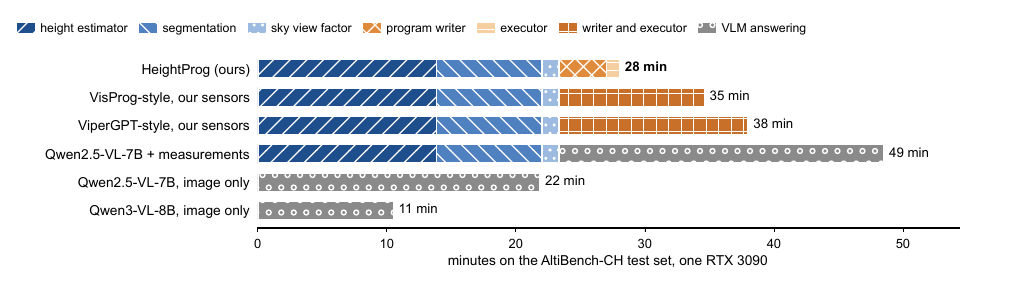}
\caption{\textbf{Run time on the \bench{}-CH test set.} Systems that read our layers start from the same sensor
run. For the measurement reader, the last segment includes writing the measurements as text.}
\label{fig:cost}
\end{figure}

\noindent\textbf{Run time.}
We time every local system on the \bench{}-CH test set, one after another on one RTX 3090 and including model
loading (Fig.~\ref{fig:cost}). \method{} takes 28 minutes, 23 of them in its sensors. VisProg- and ViperGPT-style programs over the same
sensors take 35 and 38 minutes, mainly because more of their programs go through repair rounds. Qwen2.5-VL-7B reading
our measurements takes 49 minutes, and the image-only VLMs take 11 to 22 minutes.

\end{document}